%% file: neurips_2026.tex
\documentclass{article}

\usepackage[preprint]{neurips_2026}

\usepackage[utf8]{inputenc} 
\usepackage[T1]{fontenc}    
\usepackage{url}            
\usepackage{booktabs}       
\usepackage{amsfonts}       
\usepackage{nicefrac}       
\usepackage{microtype}      
\usepackage{xcolor}         
\usepackage[most]{tcolorbox}
\usepackage{graphicx}
\usepackage{amsmath}
\usepackage{enumitem}
\usepackage{subfig}
\usepackage{booktabs} 
\usepackage{listings} 
\usepackage{float}
\usepackage{hyperref}       
\usepackage{cleveref}

\title{TIER: Trajectory-Invariant Execution Rewards for Multi-Step Tool Composition}

\author{%
  Anay Kulkarni\thanks{Equal contribution.} \\
  UC San Diego \\
  \texttt{ankulkarni@ucsd.edu}
  \And
  ChiaEn Lu\footnotemark[1] \\
  UC San Diego \\
  \texttt{allenlu@ucsd.edu}
  \And
  Dheeraj Mekala \\
  UC San Diego \\
  \texttt{dmekala@ucsd.edu}
  \AND
  Jayanth Srinivasa \\
  Cisco Research \\
  \texttt{jasriniv@cisco.com}
  \And
  Gaowen Liu \\
  Cisco Research \\
  \texttt{gaoliu@cisco.com}
  \And
  Jingbo Shang\thanks{Corresponding author.} \\
  UC San Diego \\
  \texttt{jshang@ucsd.edu}
}

\begin{document}

\maketitle

\begin{abstract}
  \input{0-abs}

\end{abstract}
\vspace{-1em}
\input{1-intro}
\vspace{-1em}
\input{2-tier}
\vspace{-0.8em}
\input{3-depthbench}
\vspace{-0.8em}
\input{4-results}
\vspace{-0.8em}
\input{5-relatedwork}
\vspace{-0.8em}
\input{6-conclusion}

\vspace{-0.8em}

\input{7-limitations}
\vspace{-0.8em}
\input{8-impact}


\bibliographystyle{plainnat}
\bibliography{references}

\input{appendix}


\end{document}

%% file: 0-abs.tex
Tool use enables large language models to solve complex tasks through sequences of API calls, yet existing reinforcement learning approaches fail to scale to multi-step composition settings.
Outcome-based rewards provide only sparse feedback, while trajectory-supervised rewards depend on annotated reference solutions, penalizing valid alternatives and limiting scalability. 
We propose \textbf{TIER}: Trajectory-Invariant Execution Rewards, a reward framework that derives supervision directly from function schemas and runtime execution, rather than from reference trajectories.
The reward decomposes into format validity, schema adherence, execution success, and answer correctness, providing dense, interpretable sequence-level feedback derived from fine-grained verification of individual steps of tool use. 
This design allows any valid execution path to receive credit, naturally supporting multiple solution strategies and adapting to evolving tool interfaces.
On DepthBench, a compositional benchmark stratified by depth (1 to 6 steps), TIER achieves $>$90\% accuracy across steps, where trajectory-supervised rewards collapse beyond step-4. 
We further demonstrate consistent gains on benchmarks like BFCL v3 and NestFUL. 
Ablation studies confirm that all reward components are necessary, highlighting the importance of multi-level supervision for compositional reasoning. 
Code and datasets are available at \url{https://github.com/anaykulkarni/TIER}

%% file: 1-intro.tex
\begin{figure*}[t]
  \centering

  \begin{minipage}[t]{0.5\textwidth}
    \vspace{3em}
    \centering
    \includegraphics[width=\linewidth]{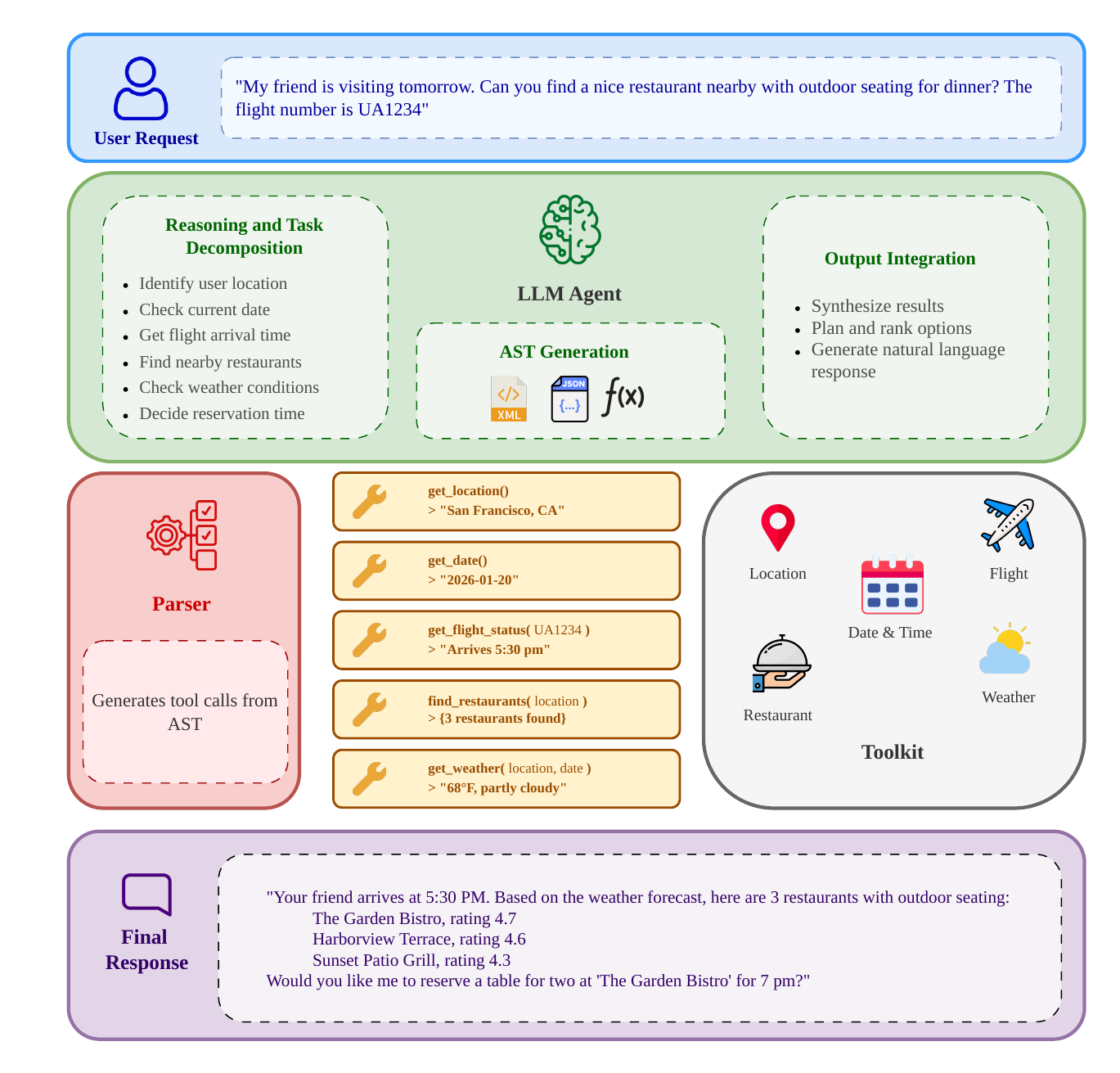}
  \end{minipage}%
  \hfill
  \begin{minipage}[t]{0.49\textwidth}
    \vspace{0pt}
    \centering
    \includegraphics[width=\linewidth]{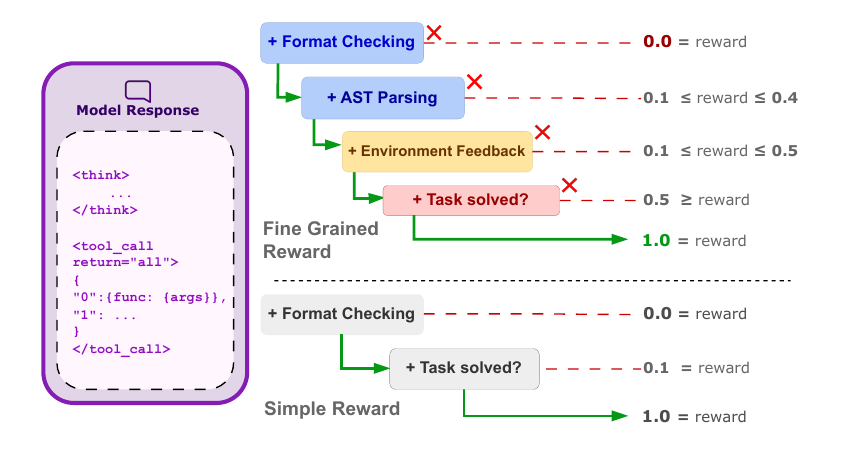}

    \vspace{0.5em}

    \includegraphics[width=\linewidth]{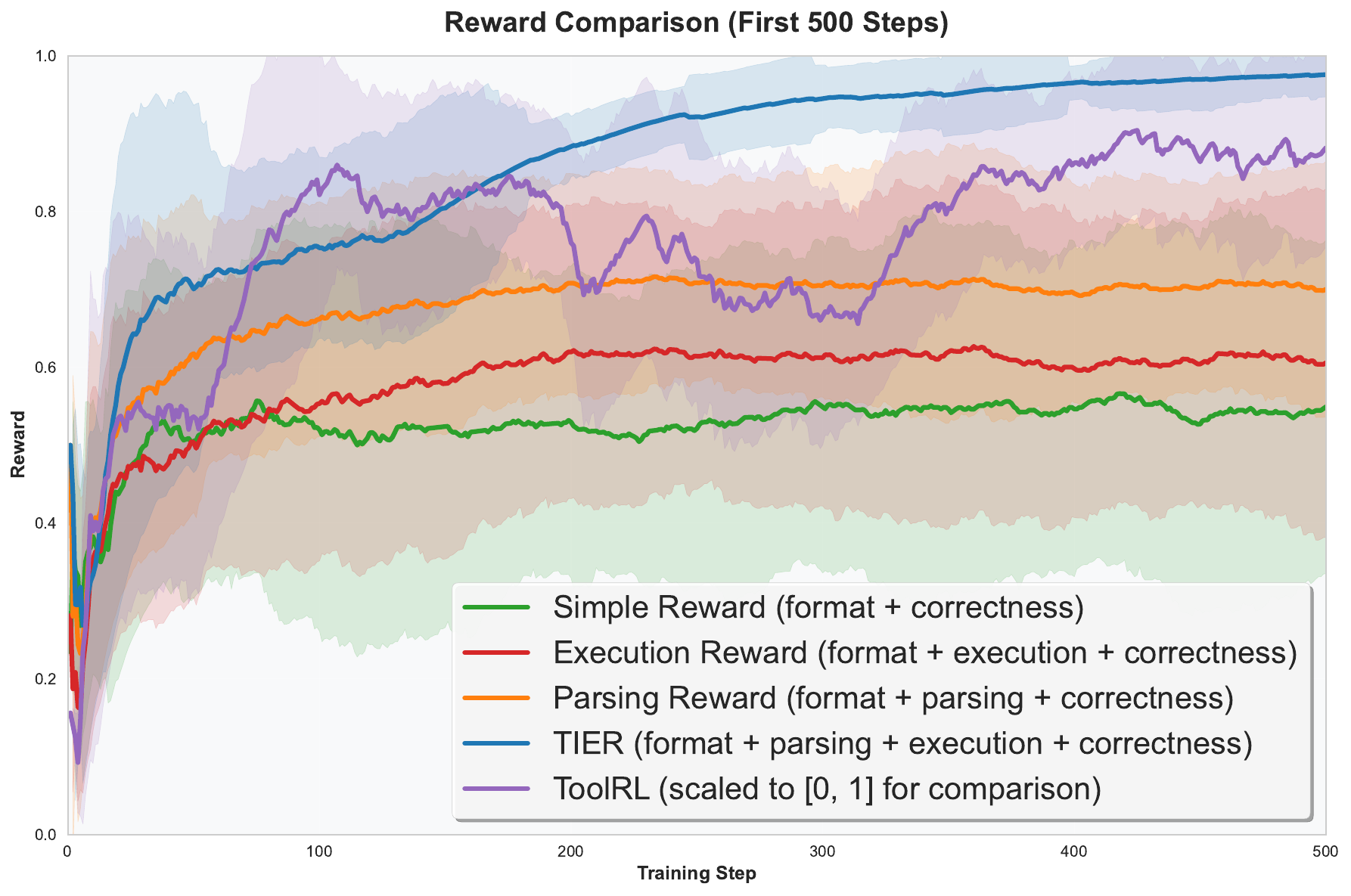}
  \end{minipage}

  \caption{\textbf{Left:} Overview of TIER for tool-augmented LLM agents. The model generates structured tool calls as abstract syntax trees (ASTs), which are executed against external tools. Unlike trajectory-supervised approaches, rewards are computed directly from function schemas and execution outcomes. \textbf{Right (Top):} TIER provides fine-grained, multi-level supervision decomposing into format, schema, execution, and answer correctness. In contrast, simple outcome rewards offer only sparse binary feedback. \textbf{Right (Bottom):} Training curves under different reward designs. Only the full fine-grained TIER reward achieves stable convergence to high performance. Partial rewards plateau early, and trajectory-supervised rewards (ToolRL) exhibit high variance due to inconsistent supervision across alternative valid compositions. All training reward curves are exponentially weighted moving averages of per-step batch-mean reward logged over a 50-step rolling window (smoothing constant, $\alpha=0.01$). The shaded region is a $\pm1$ rolling standard deviation. For visualization, we linearly rescale the rewards to $[0,1]$}
  \label{fig:tier_combined}
  \vspace{-0.3in}
\end{figure*}

\section{Introduction}

Large language models (LLMs) have shown strong capabilities in tool use, enabling interaction with external environments through structured API calls~\citep{schick2023toolformer,qin2023toolllm}.
Recent reinforcement learning (RL) approaches further improve tool-use performance~\citep{qian25,Guo25}, but they primarily target single-step tool invocation or aggregate success metrics.
In contrast, tasks that require composing multiple tool calls remain challenging, with performance degrading rapidly as composition depth increases.

Tool composition refers to tasks where solutions require multiple interdependent tool calls. For example, answering \textit{"What is the weather at the destination of my next flight?"} requires first retrieving the flight information, extracting its destination, and only then querying the weather. We refer to the longest chain of tools required for a task as its \textit{depth}. We demonstrate that under simple outcome rewards, accuracy can drop from over 94\% on 1-step, to below 2\% at 2-step, and even 0\% beyond 4-step compositions (\Cref{tab:reward_comparison_composition}). This limits the deployment of LLM-based agents in real-world settings.

A key challenge lies in reward design. Outcome-based reinforcement learning with verifiable rewards (RLVR) provides only a sparse, binary signal upon task completion, making credit assignment for intermediate decisions such as tool selection, argument specification, and execution order, difficult~\citep{shao2024deepseekmath,Guo25}.
Recent works introduce more fine-grained supervision by comparing model outputs to human-curated reference trajectories~\citep{qian25, zeng2025tool}. 
However, trajectory-based rewards penalize valid alternative solutions, require costly annotation, and are vulnerable to changes in tool interfaces.
These limits compound in compositional settings, where often multiple valid execution paths exist.

We propose TIER (\Cref{fig:tier_combined}), a reward framework for tool learning that derives supervision directly from function schemas and runtime execution rather than reference trajectories. Each tool call is evaluated through a structured process that verifies format validity, schema adherence, execution success, and answer correctness, yielding dense, interpretable feedback at the sequence level. Because the verifier reads schemas directly at training time, supervision tracks schema changes automatically; unlike trajectory-supervised approaches~\citep{qian25}, no reference trajectories must be regenerated when tools are added, modified, or deprecated. Any valid solution path receives appropriate credit, admitting multiple correct compositions that are typical of multi-step tasks. Empirically, TIER exceeds 90\% accuracy across all composition depths on DepthBench and improves on external benchmarks including BFCL v3 and NestFUL.

Our contributions are as follows:
\begin{itemize}[noitemsep, topsep=0pt, leftmargin=*]
    \item We introduce TIER, a trajectory-invariant reward framework that leverages function schemas and execution feedback, eliminating the need for annotated reference trajectories.
    \item We introduce DepthBench, a depth-stratified benchmark spanning compositions of 1 to 6 steps, enabling systematic evaluation of compositional generalization.
    \item We show that reward components are complementary and not exchangeable, emphasizing that graded, multi-level supervision is essential for higher composition depths. We support our claims with gains on DepthBench and external benchmarks (BFCL v3, NestFUL).
\end{itemize}

We publish DepthBench, the training data, and code, to support reproducibility and facilitate future research on compositional tool use: \url{https://github.com/anaykulkarni/TIER} 

%% file: 2-tier.tex
\section{TIER: Trajectory-Invariant Execution Rewards}
\label{sec:method}

TIER is built on a simple principle: rewards for tool-use trajectories should depend on \emph{what the sequence accomplishes}, not \emph{which path it follows}. Existing approaches only partially satisfy this requirement. Simple outcome-based rewards are trajectory-invariant, but provide no feedback on intermediate decisions, making credit assignment increasingly difficult as composition depth grows. In contrast, trajectory-supervised rewards offer graded feedback, but anchor it to a single reference trajectory, penalizing alternative yet valid solutions---a limitation that becomes more severe as the number of valid paths increases with depth. TIER bridges this gap by providing fine-grained, execution-grounded feedback derived directly from function schemas and runtime behavior, without relying on reference trajectories. The remainder of this section formalizes the notion of trajectory invariance and its implications, before introducing the four reward components that enable it.

\subsection{Design principle: trajectory invariance}

A reward function $R$ is \emph{trajectory-invariant} if its value on a candidate sequence $y$ depends only on properties verifiable from $y$ and the tool environment—format validity, schema adherence, execution success, and answer correctness—and not on $y$'s similarity to any reference trajectory $y^*$. 


This invariance has two practical consequences. On the space of valid solutions, all sequences satisfying the verifiable success, receive the maximum reward, regardless of which path was taken: alternative orderings of independent sub-goals, commutative operations, and functionally equivalent APIs all receive equal credit. On the space of invalid solutions, sequences are differentiated only by which verifiable properties they violate, providing a graded learning signal that does not depend on a reference trajectory. Trajectory-supervised rewards lack the first property—they arbitrarily privilege one valid path—while outcome-based rewards lack the second, collapsing the diversity of all error states into one.

A second benefit is operational: because rewards are derived from schemas and runtime execution, supervision tracks tool changes automatically, with no need to regenerate reference data when tools are added, modified, or deprecated.

TIER enforces trajectory invariance by construction: all reward components are evaluated against the current function schema and runtime behavior, with no reference to ground-truth trajectories. As shown in \Cref{sec:need_tier}, violating this property leads to failure at higher composition depths: trajectory-supervised rewards collapse precisely because they penalize valid alternative solutions, while outcome-based rewards fail due to insufficient feedback.

\subsection{Reward definition}
\label{subsec:reward_definition}

TIER reward is computed once per generated tool-call sequence as
\[
R_{\text{total}} = R_{\text{format}} + R_{\text{parse}} + R_{\text{exec}} + R_{\text{answer}}, \quad \text{with } R_{\text{total}} \text{ scaled to } [0, 1]
\]
Each component verifies a distinct property of the full sequence—syntactic, structural, operational, and semantic—and is computed without reference to ground-truth trajectories.
We represent tool sequences as structured ASTs that expose call ordering, argument bindings, and any nesting structure.


\textbf{Format Validity.}
$R_{\text{format}} \in \{0,1\}$ checks whether the full AST is well-formed and parsable. If this fails, all downstream rewards are zero.

\textbf{Schema Adherence (Parsing).} Decomposes as
\[
R_{\text{parse}} = R_{name} + R_{param} + R_{dtype}, \quad R_{\text{parse}} \in [0,3]
\]
reflecting three granularities of schema correctness. Tool names are checked categorically
\[
R_{name} =
\begin{cases}
1 & \text{if all tool names are valid} \\
0 & \text{otherwise}
\end{cases}
\]
since name validity determines which schema each call is checked against, any invalid API makes the sequence unverifiable and by extension invalid ($R_{\text{parse}} = 0$). Parameter and type correctness are graded by mismatch counts,
\[
R_{param/dtype} = \text{clip}(1 - \lambda_p \cdot p, 0, 1),
\]
where $p$ is the total number of mismatches across all calls in the sequence, each call's parameters checked against its own schema. We set $\lambda_p = 0.25$, balancing two failure modes: too large a coefficient eliminates partial credit for nearly-correct schemas, while too small a coefficient makes individual mismatches negligible.

\textbf{Execution Success}
$R_{\text{exec}} \in \{0,1\}$. Equals 1 iff every tool call executes successfully; otherwise 0. The all-or-nothing rule reflects the structure of compositional dependence: a failure at step $k$ invalidates downstream calls that depend on its output, so a graded reward (e.g., fraction of successful calls) would credit work that produced no usable result. Treating chain-breaking failures as fully penalizing aligns the reward with the actual structure of compositional success.

\textbf{Answer Correctness.}
$R_{\text{answer}} = 5 \cdot1[\text{answer correct}] \in \{0, 5\}$. Equals 5 iff the final output matches the expected answer. We weight answer correctness substantially higher than the other components because lower weightings induce reward hacking — the model satisfies format, schema, and execution validity without producing correct answers, and training stalls before the answer correctness signal can take effect. The weighting prioritizes end-task success while preserving graded shaping signal from the lower-level components.

\textbf{Sequence-level aggregation and trajectory invariance.} All four components are evaluated over the full sequence rather than accumulated across per-step. Two consequences follow. First, all sequences producing the correct final answer and satisfying intermediate validity receive the maximum reward, regardless of length, ordering of independent sub-goals, or choice among functionally equivalent APIs. Second, sequences that fail are differentiated by which verifiable properties they violate providing graded correction signal that does not depend on any reference trajectory. TIER is therefore trajectory-invariant on the space of valid solutions, and informative on the space of invalid ones.

\subsection{Reinforcement learning}
\label{subsec:rl}

We train models using a GRPO-style policy gradient objective with group-normalized advantages and token-level loss aggregation. Given a prompt $x$, we sample a set of trajectories $\mathcal{G}(x) = \{y_1, \dots, y_K\}$ from the current policy $y \sim \pi_\theta(\cdot \mid x)$ and execute each to obtain scalar rewards $R(x,y) \in [0,1]$.

\textbf{Group-normalized advantages.}
Advantages are computed at the trajectory level by normalizing rewards within each rollout group:
\[
A(x,y) = \frac{R(x,y) - \mu_{\mathcal{G}(x)}}{\sigma_{\mathcal{G}(x)} + \varepsilon},
\]
where $\mu_{\mathcal{G}(x)}$ and $\sigma_{\mathcal{G}(x)}$ are the mean and standard deviation of rewards over $\mathcal{G}(x)$, and $\varepsilon = 10^{-4}$.

\textbf{Policy gradient objective.}
We optimize a PPO-style clipped surrogate objective at the token level. For a trajectory $y = (y_1, \dots, y_T)$:
\[
\rho_t(\theta) = \frac{\pi_\theta(y_t \mid x, y_{<t})}{\pi_{\theta_{\text{old}}}(y_t \mid x, y_{<t})},
\]
\[
\mathcal{L}_{\text{PG}}(\theta) = 
- \frac{1}{\sum_{i \in \mathcal{B}} T_i}
\sum_{(x,y) \in \mathcal{B}} 
\sum_{t=1}^{T_y}
\min\!\Big(
\rho_t(\theta) A(x,y),\;
\mathrm{clip}(\rho_t(\theta), 1-\epsilon, 1+\epsilon)\, A(x,y)
\Big),
\]
where $\mathcal{B}$ denotes the local batch and $T_y$ is the number of active (non-padding) tokens in trajectory $y$. The loss is normalized by the total number of active tokens in the batch.

In our setup, we use a single update per rollout ($\texttt{num\_iterations}=1$), so $\pi_{\theta_{\text{old}}} = \pi_\theta$ at the update step and $\rho_t(\theta)=1$. In this case, the clipped surrogate reduces in gradient to the standard REINFORCE estimator \citep{williams1992simple} $-A(x,y)\nabla_\theta \log \pi_\theta(y_t \mid x, y_{<t})$.

\textbf{KL regularization.}
We apply per-token KL regularization to a reference policy $\pi_{\text{ref}}$ using the k3 estimator \citep{schulman2017kl}:
\[
\hat{\mathrm{KL}}_t =
\exp\!\big(\log \pi_{\text{ref}}(y_t \mid \cdot) - \log \pi_\theta(y_t \mid \cdot)\big)
- \big(\log \pi_{\text{ref}}(y_t \mid \cdot) - \log \pi_\theta(y_t \mid \cdot)\big)
- 1,
\]
\[
\mathcal{L}(\theta) =
\mathcal{L}_{\text{PG}}(\theta)
+ \frac{\lambda_{\mathrm{KL}}}{\sum_{i \in \mathcal{B}} T_i}
\sum_{(x,y) \in \mathcal{B}} \sum_{t=1}^{T_y} \hat{\mathrm{KL}}_t, \quad \text{with } \lambda_{\mathrm{KL}} = 0.04
\]

The reward formulation defined in \Cref{subsec:reward_definition} is agnostic to the optimization algorithm. We verify this by evaluating three policy optimization variants: GRPO-style, batch-normalized variants, and DAPO \citep{yu2025dapo} under TIER, observing comparable performance (see \Cref{appendix:algorithm_ablation}).

%% file: 3-depthbench.tex
\section{DepthBench: A depth-stratified benchmark for multi-step tool composition}
\label{sec:environment_dataset}

\subsection{Environment}

DepthBench evaluates multi-step tool composition over a fully executable backend of 163 hand-crafted tools spanning three categories: information and utility functions, structured retrieval over synthetic datasets, and multi-step workflows. All tools execute deterministically, supporting both reproducible evaluation and the runtime grading needed to compute TIER's execution and answer rewards. A rule-based parser validates schema adherence on each call, serving as the evaluation harness at test time and the $R_\text{parse}$ verifier during training. To evaluate transfer beyond this controlled setting, we additionally benchmark on BFCL v3~\citep{bfcl25} and NestFUL~\citep{basu25}

\subsection{DepthBench design and evaluation protocol}

DepthBench contains 1,710 queries stratified by composition depth from 0 to 6, where depth 0 denotes irrelevance cases in which no available tool can answer the query, and depths 1–6 denote single-call through six-call compositions. Multi-call compositions (depths 2–6) account for roughly half of the benchmark; the remainder consists of single-call queries and irrelevance cases. Composition structures include linear chains ($A \rightarrow B \rightarrow C$), funnels ($(A,B)\rightarrow C$), trees ($A\rightarrow(B,C)$), and order-agnostic conjunctive structures ($A \rightarrow (B \& C \& D) \rightarrow E$). 

Although DepthBench is smaller than prior tool-use benchmarks, its construction is deliberately controlled: tool schemas are modeled after commonly-used APIs with executable backends implemented to simulate their behavior; queries are hand-crafted and verified against their intended composition structure. Comparable benchmarks make different trade-offs — BFCL v3 is LLM-generated and lacks an executable backend, while NestFUL provides 1,861 evaluation examples but is not stratified by depth — leaving DepthBench as the only setting we are aware of that combines realistic, executable tool semantics with depth-stratified evaluation. This stratification, absent from prior work(~\citep{bfcl25, basu25, zhong25}), enables systematic analysis of compositional generalization rather than aggregate success.

Train–test splits use disjoint sets of target tools, with distractors also disjoint between splits, ensuring that no tool seen at training time appears at evaluation time in any role. Distractors are deliberately chosen to share schemas and output types with target tools, requiring models to distinguish functionally similar APIs rather than relying on superficial cues such as names or signature patterns. For multi-step evaluation, the dataset is split into 944 training and 766 validation examples, with tool schemas and sampled distractors provided at both training and evaluation time.

%% file: 4-results.tex
\section{Experiment Results}
\label{sec:experiment_results}

\subsection{Experimental setup}

\textbf{Model and Training.}
All experiments use Qwen3-8B as the base model. We train on a context length of 12,288 tokens with batch-normalized GRPO objective \citep{shao2024deepseekmath}, regularized against a fixed reference policy (KL penalty $\lambda_{\text{KL}}=0.04$. For each prompt, we sample $|\mathcal{G}|=8$ trajectories and compute group-normalized advantages \Cref{subsec:rl}. Unless stated otherwise, the reward formulation from \Cref{subsec:reward_definition} is held fixed; \Cref{subsec:reward_ablation} provides ablations over individual reward components. To isolate the effect of reward design from optimization, we additionally evaluate other RL objectives (GRPO, DAPO) under the same reward in \Cref{appendix:algorithm_ablation}.

\textbf{Baselines.}
We compare TIER against representative alternatives: (1) \textit{Simple (outcome-based)}, a binary reward based only on format validity and final answer correctness, with no intermediate supervision; (2) \textit{ToolRL (trajectory-supervised)} \citep{qian25}, which uses fine-grained rewards derived from matching reference trajectories; (3) \textit{Qwen3-8B (zero-shot)}, the pretrained base model; (4) \textit{Qwen3-8B (3-shot ICL)}, the base model with three in-context examples; and (5) \textit{GPT-5 (3-shot ICL)}, a frontier reference under the same 3-shot protocol, included as a scale reference rather than a head-to-head comparison given its substantially larger parameter count. All RL variants share the same model, optimizer, rollout budget, and hyperparameters; only the reward differs.

\textbf{Evaluation Benchmarks.}
We evaluate on three benchmarks: (1) \textit{DepthBench} (\Cref{sec:environment_dataset}), is evaluated on the 766-example validation set, with results stratified by composition depth (0–6). (2) \textit{BFCL v3} \citep{bfcl25}, a real-world function-calling benchmark with different tools and evaluation protocols that measures out-of-distribution generalization; and (3) \textit{NestFUL} \citep{basu25}, which evaluates nested and sequential API composition, where we report a controlled comparison against ToolRL under Qwen3-8B, and a broader comparison with tool-use models trained by other mechanisms. Finally, to evaluate whether DepthBench's compact, composition-focused supervision is competitive with substantially larger tool-use corpora, we additionally train Qwen3-8B under TIER on RL adapted versions of ToolACE~\citep{toolace} and xLAM~\citep{xlam} (\Cref{appendix:dataset_analysis}).

\subsection{Multi-step composition requires trajectory invariance and execution awareness}
\label{sec:need_tier}

\begin{figure*}[t]
  \centering
  \label{fig:performance_comparison_fig}
  \subfloat{
    \includegraphics[width=0.49\textwidth]{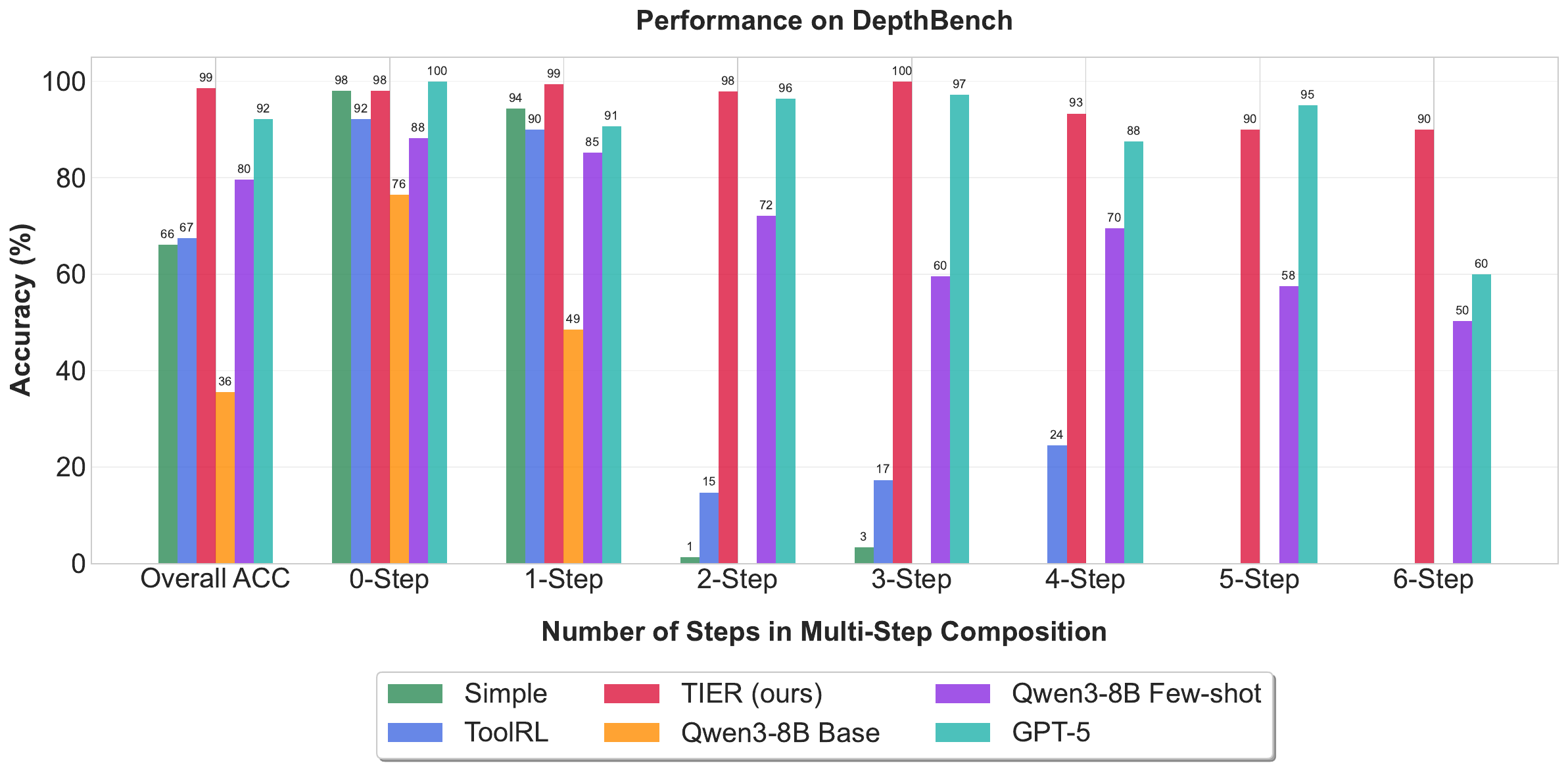}
    \label{fig:performance_ours}
  }
  \subfloat{
    \includegraphics[width=0.49\textwidth]{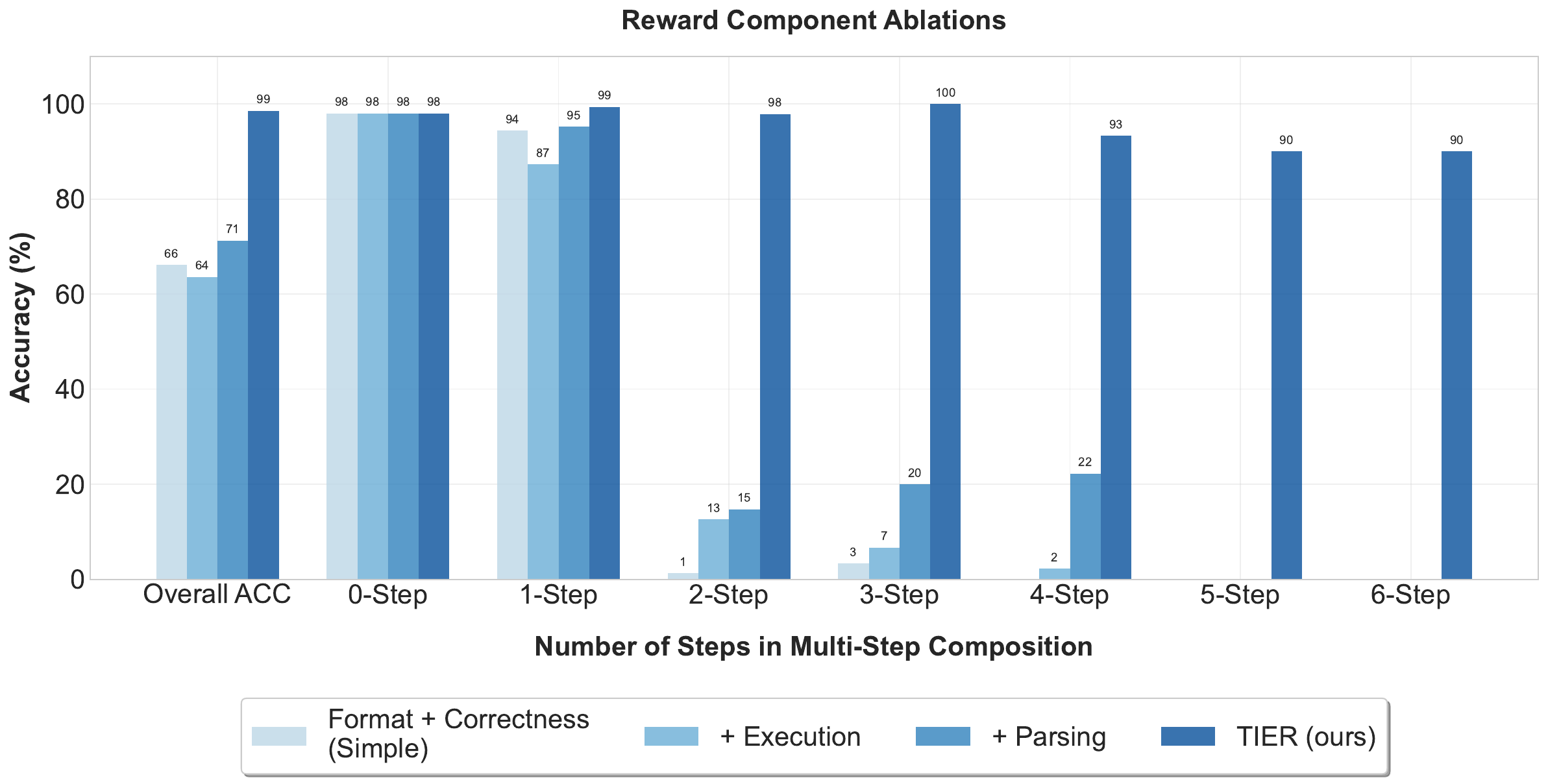}
    \label{fig:performance_reward_ablation}
  }
  \caption{\textbf{(a)} Performance on DepthBench. \textbf{(b)} Reward component ablations under TIER 
  }
  \label{fig:reward_ablation}
  \vspace{-1em}
\end{figure*}

We test the central claim that multi-step composition emerges only under TIER. We report DepthBench accuracy stratified by composition depth along with population-weighted average of per-depth accuracies in \Cref{tab:reward_comparison_composition}, \Cref{fig:performance_ours}. 

\begin{table*}[t]
  \centering
  \caption{Performance comparison on DepthBench, stratified by composition depth.}
  \label{tab:reward_comparison_composition}
  \small
  \setlength{\tabcolsep}{4pt}
  \resizebox{\textwidth}{!}{
    \begin{tabular}{lccccccccc}
      \toprule
      Reward Model & Method & Overall Acc & 0-Step & 1-Step & 2-Step & 3-Step & 4-Step & 5-Step & 6-Step \\
      \midrule
      Simple & RL & 66.15 & 98.04 & 94.38 & 1.25 & 3.33 & 0.00 & 0.00 & 0.00 \\
      ToolRL & RL & 67.49 & 92.16 & 90.00 & 14.74 & 17.33 & 24.44 & 0.00 & 0.00 \\
      TIER (ours) & RL & \textbf{98.57} & \textbf{98.04} & \textbf{99.38} & \textbf{97.89} & \textbf{100.00} & \textbf{93.33} & 90.00 & \textbf{90.00} \\
      \midrule
      Qwen3-8B & Zero-shot & 35.51 & 76.47 & 48.54 & 0.00 & 0.00 & 0.00 & 0.00 & 0.00 \\
      Qwen3-8B & 3-shot ICL & 79.56 & 88.24 & 85.25 & 72.12 & 59.55 & 69.57 & 57.50 & 50.25 \\
      GPT-5 & 3-shot ICL & 92.16 & 98.02 & 90.11 & 96.36 & 97.28 & 87.59 & \textbf{95.00} & 60.00 \\
      \bottomrule
    \end{tabular}
  }
  \vspace{-1em}
\end{table*}

Three observations follow. 
(1) TIER sustains high performance across all depths achieving $\geq$90\% accuracy at every composition depth, including 90\% at depths 5 and 6 — the regime where depth-6 tasks introduce compound structures combining funnels and trees that no shallower depth contains.

(2) Outcome-based rewards fail beyond shallow composition. Simple-RL drops from 94.38\% at one step to 1.25\% at two steps and zero beyond four steps. Sparse binary signals provide no guidance for invalid solutions. Missing diversity in error states captured by the reward makes training over multi-step trajectories intractable. Notably this produces high-variance reward signals (\Cref{fig:tier_combined})

(3) Trajectory supervision collapses for a different reason. Although ToolRL provides intermediate feedback, it penalizes valid alternative compositions, reaching 0\% accuracy at depths 5 and 6 where multiple valid paths are common. Training dynamics (\Cref{fig:tier_combined}) clarify the mechanism: ToolRL's reward exhibits high-variance oscillations because correct executions deviating from reference trajectories receive low reward, pushing the policy away from valid solutions. Notably the failure stems from a misspecified reward, not insufficient training.

RL versus prompting. Naive 3-shot ICL on Qwen3-8B (79.56\%) outperforms both Simple-RL (66.15\%) and ToolRL (67.49\%), indicating that poorly-designed RL rewards can underperform no RL at all. ICL provides explicit per-structure demonstrations that bias the model toward valid compositions; in the absence of useful reward signal, RL training on Simple or ToolRL erodes this prior rather than building on it. TIER, by contrast, supplies reward signal aligned with composition structure, recovering and exceeding ICL's gains across all depths.

\subsection{Reward-component ablation}
\label{subsec:reward_ablation}
While \Cref{tab:reward_comparison_composition} establishes that TIER enables multi-step composition, it does not isolate which components are necessary. We ablate the reward by removing components individually, holding the base model, optimizer, rollout budget, and RL algorithm fixed. Format and answer correctness are retained in all settings as the minimal baseline (identical to Simple-RL of \Cref{tab:reward_comparison_composition}). When a component is removed, its weight is redistributed and the total renormalized to $[0, 1]$ to control for scale. Results are shown in \Cref{tab:reward_granularity_comparison}, \Cref{fig:performance_reward_ablation}.

\begin{table*}[t]
  \centering
  \caption{Effects of reward granularity as seen on validation set}
  \label{tab:reward_granularity_comparison}
  \small
  \setlength{\tabcolsep}{4pt}
  \resizebox{\textwidth}{!}{
    \begin{tabular}{lcccccccc}
      \toprule
      Reward Model & Overall Acc & 0-Step & 1-Step & 2-Step & 3-Step & 4-Step & 5-Step & 6-Step \\
      \midrule
      Format + Correctness (Simple) & 66.15 & 98.04 & 94.38 & 1.25 & 3.33 & 0.00 & 0.00 & 0.00 \\
      + Execution (no Parsing) & 63.58 & 98.04 & 87.29 & 12.63 & 6.67 & 2.22 & 0.00 & 0.00 \\
      + Parsing (no Execution) & 71.28 & 98.04 & 95.21 & 14.74 & 20.00 & 22.22 & 0.00 & 0.00 \\
      TIER (full reward) & \textbf{98.57} & 98.04 & 99.38 & 97.89 & 100.00 & 93.33 & 90.00 & 90.00 \\
      \bottomrule
    \end{tabular}
  }
  \vspace{-1em}
\end{table*}

\textbf{A sharp sufficiency threshold.}
Partial rewards fail abruptly beyond shallow depths. Simple drops to 0\% beyond four steps; adding either execution or parsing yields modest improvement at intermediate depths but still collapses to 0\% at depths 5 and 6. Only the full reward sustains $\geq$90\% across all depths. The transition from 0\% to 90\% at six steps under identical training represents a qualitative shift in training outcomes rather than gradual improvement: the components are not substitutes whose contributions sum incrementally, but a set whose union unlocks compositional behavior.

A counterintuitive result: adding execution reward without parsing decreases overall accuracy, with 1-step accuracy also dropping. We inspect failures by reviewing rollout logs at 100-step intervals throughout training. We observe a reward-hacking pattern: in the absence of schema-level discrimination from parsing, the model learns to route most queries to a small set of "safe" tools that reliably execute, accumulating execution reward while answering incorrectly. The weighting of answer correctness over execution prevents this collapse only when parsing supplies discrimination signal, but is insufficient on its own. This supports the rationale of designing parsing and execution as complementary components: parsing enforces structural validity (correct tool signature) and execution enforces functional validity (no runtime error). Neither suffices alone.

\textbf{Failure is due to signal saturation, not instability.}
(\Cref{fig:tier_combined}, right) show that all partial-reward configurations converge stably but plateau below the optimum. The model saturates the available signal; the signal itself lacks the structure needed for compositional behavior. Only the full reward reaches near-ceiling performance.


\textbf{Implications.}
 Multi-step composition requires concurrent supervision at three levels: syntactic (format), structural (schema), and functional (execution). Errors at each level produce different failure modes: format errors block parsing entirely, schema errors propagate to downstream calls, semantic errors corrupt outputs without raising syntactic flags. The fully decomposed reward, however, addresses error states at their origin. Compositional tool learning is therefore not a single objective but a multi-level credit-assignment problem requiring reward signals targeted at distinct failure modes.

\subsection{Transfer to external benchmarks}
\label{subsec: external_benchmarks}
\begin{table*}[t]
  \centering
  \caption{Comparison across different training objectives on BFCL.}
  \label{tab:reward_comparison_bfcl}
  \small
  \setlength{\tabcolsep}{5pt}
  \resizebox{\textwidth}{!}{
    \begin{tabular}{lcccccc}
      \toprule
      Reward format & Overall & Non-Live AST & Live & Multiturn & Relevance & Irrelevance \\
      \midrule
      TIER & \textbf{68.92} & 85.94 & \textbf{81.96} & \textbf{39.12} & 77.78 & \textbf{87.65} \\
      Simple & 66.30 & 85.00 & 78.72 & 34.00 & 77.78 & 83.56 \\
      ToolRL & 37.27 & 84.10 & 43.98 & 0.38 & \textbf{94.44} & 0.81 \\
      Supervised Fine Tuning & 61.47 & 84.12 & 79.52 & 19.25 & 61.11 & 90.33 \\
      Base (Qwen3-8B) & 64.31 & \textbf{88.81} & 78.54 & 33.38 & 77.78 & 79.08 \\
      \bottomrule
    \end{tabular}
  }
  \vspace{-1.5em}
\end{table*}

\textbf{BFCL.}
We evaluate generalization beyond the training environment on BFCL v3. All models are trained on a 1,240 example split from DepthBench and differ only in reward formulation. Additionally, we include a supervised fine-tuning (SFT) baseline and the pretrained base model for reference. To support evaluation on BFCL's long-context tasks, we extend our model's context with YaRN \citep{peng2023yarn}. As shown in \Cref{tab:reward_comparison_bfcl}, TIER achieves the best overall performance (68.92\%) and outperforms all baselines. ToolRL reaches comparable Non-Live AST accuracy (84.10), but collapses on Multiturn (0.38) and Irrelevance (0.81), where success requires adaptive decisions rather than reproducing a fixed trajectory. Simple rewards provide modest gains over the base model but remain below TIER, while SFT performs worse overall (61.47\%) and degrades relative to the base model, reflecting overfitting to single reference trajectories. These results show that TIER rewards transfer better to real-world function-calling settings balancing both correctness and adaptability.

\textbf{NestFUL.}
We evaluate on NestFUL under two protocols. In a controlled zero-shot comparison (\Cref{tab:nestful_combined}(a)), TIER achieves a 0.684 exact-match accuracy versus ToolRL’s 0.476, confirming the advantage of TIER rewards. Under 3-shot ICL (\Cref{tab:nestful_combined}(b)), our post-trained 8B model achieves 0.75, outperforming substantially larger models such as DeepSeek-V3 (685B, 0.60) and GPT-4o (0.60). In contrast, SFT-based models such as xLAM and Hammer2.1 achieve near-zero accuracies despite strong single-call accuracy, reinforcing that SFT alone is insufficient for compositional reasoning.

\begin{table*}[t]
  \centering
  \small
  \setlength{\tabcolsep}{4pt}
  \caption{\textbf{NestFUL results.} \textbf{(a)} Training Qwen3-8B with TIER outperforms ToolRL. \textbf{(b)} TIER with Qwen3-8B compared against other tool learning methods. Exact Match Accuracy (EM Acc) is the ratio of test cases where all predicted APIs execute correctly and match the ground truth exactly. Baseline results are from~\citep{basu25}.}
  \label{tab:nestful_combined}

  \begin{minipage}[t]{0.25\textwidth}
    \vspace{0pt}
    \centering
    \textbf{(a) TIER vs. ToolRL}\\[0.4em]
    \begin{tabular}{lc}
      \toprule
      Method & EM Acc \\
      \midrule
      ToolRL & 0.476 \\
      TIER & \textbf{0.684} \\
      \bottomrule
    \end{tabular}
  \end{minipage}%
  \hfill
  \begin{minipage}[t]{0.72\textwidth}
    \vspace{0pt}
    \centering
    \textbf{(b) Comparison with tool-use models}\\[0.4em]
    \resizebox{\linewidth}{!}{
      \begin{tabular}{lccccc}
        \toprule
         & Hammer2.1-7B & TIER & xLAM-8x7b-fc-r & DeepSeek-V3 & GPT-4o \\
        \midrule
        \# Param & 7B & 8B & 46.7B & 685B & UNK \\
        EM Acc & 0.08 & \textbf{0.75} & 0.03 & 0.60 & 0.60 \\
        \bottomrule
      \end{tabular}
    }
  \end{minipage}

  \vspace{-1.5em}
\end{table*}

\subsection{Intermediate representation for tool calls}

Tool interaction requires a structured intermediate representation (IR) that can be deterministically parsed and executed. We evaluate JSON-based ASTs, XML-based ASTs, and direct generation without an IR under TIER (\Cref{tab:generation_strategy}, \Cref{fig:BFCL_ast}, \Cref{appendix:AST}). JSON achieves the best overall performance and is therefore used as the default IR in all experiments. Since structured IRs for tool use are already well established in prior work, we defer implementation details to the appendix.

%% file: 5-relatedwork.tex
\section{Related work}
\label{sec:related_work}

\textbf{Tool Use via Prompting and Supervised Learning.}
Early systems rely on prompting-based methods such as Chain-of-Thought and ReAct \citep{wei23, yao23}, as well as engineered pipelines that interleave reasoning and tool calls \citep{paranjape2023art, zhuang2023toolchain, song2023restgpt}. More recent work fine-tunes models on curated tool-use trajectories or schemas \citep{schick2023toolformer, liu2024apigen, liang2024taskmatrix}, improving robustness but requiring annotated intermediate actions and fixed tool interfaces. Several efforts train models to follow abstract function-calling schemas that generalize across tools via schema injection or retrieval \citep{qin2023toolllm, du2025generalizable, qi2024learning}, enabling plug-and-play extensibility. However, these approaches rely on supervised objectives or exact-match evaluation and do not address learning from execution feedback.

\textbf{RL for Tool Use.}
Recent work applies RL as a post-training paradigm for tool learning \citep{qian25, Guo25}. Outcome-based rewards rely solely on final task success, providing sparse feedback that does not assign credit across intermediate decisions. Concurrent work such as PORTool \citep{wu2025portool} addresses this by redistributing trajectory-level rewards using rollout-tree comparisons, but the resulting signal remains path-dependent and tied to specific exploration trajectories. ToolRL \citep{qian25} introduces fine-grained supervision, but derives it by matching model outputs to annotated reference trajectories, requiring curated data and penalizing valid alternative solutions. In contrast, TIER derives rewards directly from function schemas and runtime execution, providing fine-grained, trajectory-invariant supervision without relying on reference trajectories.

\textbf{Multi-Step Tool Composition.}
 Despite extensive work on tool learning, systematic evaluation and learning of multi-step composition remains limited. Existing benchmarks such as NestFUL \citep{basu25} and ComplexFuncBench \citep{zhong25} evaluate compositional tasks but do not stratify by composition depth. Prior RL approaches do not demonstrate reliable generalization across varying composition lengths. DepthBench addresses this gap directly by enabling per-depth accuracy measurement, which we show is essential for diagnosing reward failure modes. We demonstrate that TIER enables reliable composition where both outcome-based and trajectory-supervised rewards fail.

%% file: 6-conclusion.tex
\section{Conclusion and future work}

We show that reward design is the primary bottleneck for multi-step tool composition in RL-trained language models. Outcome-based rewards fail at deeper compositions due to sparse credit assignment, while trajectory-supervised rewards penalize valid alternative solutions, a limitation that worsens as the number of feasible execution paths grows with composition depth. TIER addresses both issues by deriving rewards deterministically from function schemas and runtime execution, enabling fine-grained supervision that is invariant to the chosen solution path. Reward-component ablations further show that compositional tool use requires supervision at the syntactic, semantic, and execution levels simultaneously; removing any component substantially degrades performance. TIER achieves $\ge$90\% accuracy across all composition depths on DepthBench and generalizes consistently to BFCL v3 and NestFUL, demonstrating strong scalability and robustness for compositional tool use.

An important direction for future work is extending TIER from single-turn to multi-turn RL. By feeding model outputs from previous turns back into subsequent turns, the model can iteratively diagnose and correct failures. Since rewards are execution-grounded, both success and failure states can be incorporated automatically into the training signal, making it well-suited for scalable multi-turn training, where constructing such datasets is otherwise prohibitive. Another promising direction is applying TIER’s underlying RLVR principle to domains such as code generation and NL-to-SQL, where correctness can also be verified deterministically, enabling RL training without trajectory supervision. Finally, the depth stratification in DepthBench suggests a natural curriculum-learning strategy: training can begin with shallow compositions and progressively introduce deeper ones as the model’s capabilities improve. Such curricula may learn deeper compositions effectively.

%% file: 7-limitations.tex
\section{Limitations}
\label{sec:limitations}

Our primary benchmark uses a synthetic, deterministic backend to enable controlled analysis of reward design, but real-world tool environments are stochastic and constrained, with latency, noisy outputs, partial failures, rate limits, monetary cost, and irreversible side effects. TIER assumes exploratory execution is cheap, repeatable, and safe, which may not hold in practice. Although BFCL v3 and NestFUL provide initial evidence of generalization, evaluating TIER in stochastic, cost-sensitive, and safety-critical settings remains future work. In addition, training is limited to a single-turn setting, while multi-turn evaluation is performed without explicit multi-turn RL training. Further we conduct all experiments at a single model scale using Qwen3-8B; although algorithm ablations in \Cref{appendix:algorithm_ablation} suggest gains primarily arise from reward design, we do not study scaling behavior across model sizes. Reward weights are fixed based on preliminary tuning, and sensitivity to different weighting schemes across domains and composition depths is not explored. Finally, due to hardware constraints, training is limited to a 12,288-token context despite Qwen3-8B supporting substantially longer contexts via YaRN \citep{qwen3, peng2023yarn}. Consequently, the model is not exposed to long-context function-calling trajectories during training, and we exclude long-context benchmarks such as ComplexFuncBench \citep{zhong25} after observing substantial degradation in preliminary experiments.

%% file: 8-impact.tex
\section{Impact Statement}
\label{sec:impact}
This work advances post-training methods for large language models with an emphasis on reward design. The proposed approach reduces dependence on costly human annotation for tool-use training, which may broaden access to capable tool-using agents. As with any advance in autonomous agent capabilities, deployment in safety-critical or high-stakes environments requires appropriate guardrails, monitoring, and human oversight. We do not identify ethical concerns specific to this work beyond those generally associated with the development and deployment of language model agents.

%% file: appendix.tex
\newpage
\appendix
\crefalias{section}{appendix}
\definecolor{darkgreen}{RGB}{0,120,0}
\definecolor{cerulean}{rgb}{0.0, 0.48, 0.65}
\definecolor{darkblue}{rgb}{0.0, 0.15, 0.4}
\definecolor{darkred}{RGB}{120, 20, 20}
\definecolor{softpeach}{RGB}{214, 140, 120}
\definecolor{softrose}{RGB}{196, 120, 120}

\newtcolorbox{astbox}{
    colback=white,
    colframe=darkgreen,
    boxrule=1pt,
    arc=3pt,
    width=\linewidth,
    title=AST,
    coltitle=white,
    colbacktitle=darkgreen,
    fonttitle=\bfseries,
    left=8pt,
    right=8pt,
    top=8pt,
    bottom=8pt
}

\newtcolorbox{promptbox}{
    colback=white,
    colframe=cerulean,
    boxrule=1pt,
    arc=3pt,
    width=\linewidth,
    title=Prompt,
    coltitle=white,
    colbacktitle=cerulean,
    fonttitle=\bfseries,
    left=8pt,
    right=8pt,
    top=8pt,
    bottom=8pt
}

\newtcolorbox{syspromptbox}{
    enhanced,
    breakable,
    colback=white,
    colframe=darkblue,
    boxrule=1pt,
    arc=3pt,
    width=\linewidth,
    title=System Prompt,
    coltitle=white,
    colbacktitle=darkblue,
    fonttitle=\bfseries,
    left=8pt,
    right=8pt,
    top=8pt,
    bottom=8pt
}

\newtcolorbox{modelbox}{
    enhanced,
    breakable,
    colback=white,
    colframe=softrose,
    boxrule=1pt,
    arc=3pt,
    width=\linewidth,
    title=Model Response,
    coltitle=white,
    colbacktitle=softrose,
    fonttitle=\bfseries,
    left=8pt,
    right=8pt,
    top=8pt,
    bottom=8pt
}

\section{AST-Based Intermediate Representations}
\label{appendix:AST}
We evaluate three strategies for function-call generation: JSON-based abstract syntax trees (ASTs), XML-based ASTs, and direct generation of tool calls without an intermediate representation.

These strategies impose different structural constraints and levels of syntactic flexibility, which affect training stability and execution correctness. 

\begin{figure}[ht]
  \begin{center}
    \centering
    \includegraphics[width=0.8 \columnwidth]{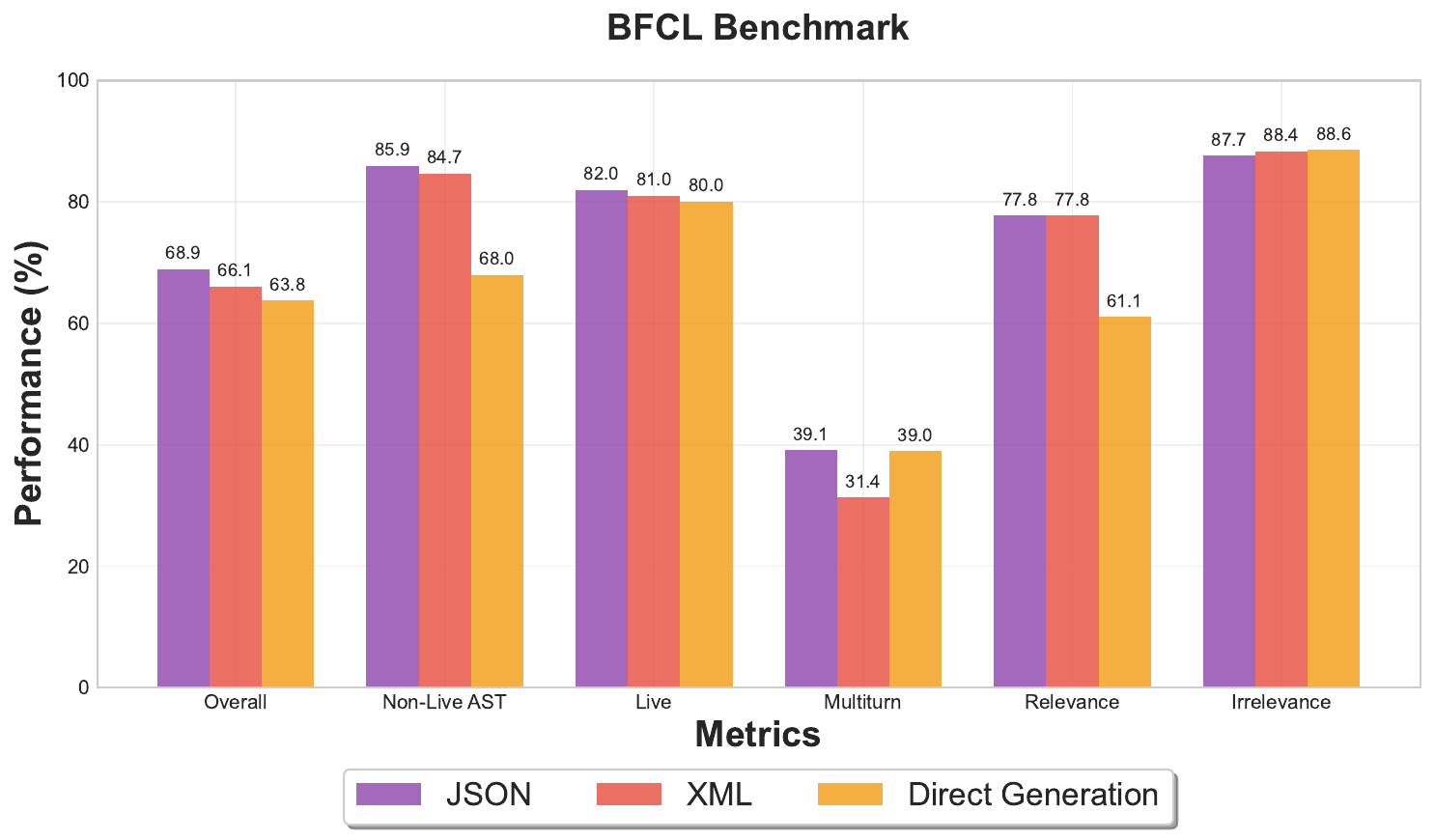}
    \caption{
      Performance of different intermediate representations under fine-grained rewards on BFCL v3.
    }
    \label{fig:BFCL_ast}
  \end{center}
 \vskip -0.3in
\end{figure}

\begin{table*}[h]
  \centering
  \caption{Comparison between generation strategies}
  \label{tab:generation_strategy}
  \small
  \setlength{\tabcolsep}{5pt}
  \resizebox{\textwidth}{!}{
    \begin{tabular}{lcccccc}
      \toprule
      IR format & Overall & Non-Live AST & Live & Multiturn & Relevance & Irrelevance \\
      \midrule
      JSON & \textbf{68.92} & \textbf{85.94} & \textbf{81.96} & \textbf{39.12} & \textbf{77.78} & 87.65 \\
      XML & 66.06 & 84.73 & 81.03 & 31.37 & 77.78 & 88.37 \\
      Direct Generation & 63.80 & 68.02 & 80.05 & 39.00 & 61.11 & \textbf{88.56} \\
      \bottomrule
    \end{tabular}
  }
\end{table*}

As shown in \Cref{tab:generation_strategy}, JSON-based IRs achieve the strongest overall performance, outperforming both XML and direct generation across most metrics, including Non-Live AST, Live, Multiturn, and overall accuracy. These results are consistent with observations by \citep{ma2024m, bfcl25}.

Direct generation performs competitively on certain subsets—matching JSON on Multiturn (39.00 vs 39.12) and slightly outperforming both structured representations on Irrelevance (88.56). However, it degrades substantially in syntactic correctness (Non-Live AST: 68.02 vs 85.94 for JSON), leading to lower overall performance.

XML improves over direct generation in syntactic correctness and overall accuracy, but this advantage is not uniform across metrics: it underperforms direct generation on Multiturn (31.37 vs 39.00) and achieves comparable but not superior performance on Irrelevance. 

Overall, structured IRs provide clear benefits for schema adherence and execution reliability, with JSON offering the best trade-off between structural consistency and downstream performance. Based on these results, we adopt JSON as the default intermediate representation in subsequent experiments.

\label{appendix:ir_specs}
\subsection{XML-Based AST Format.}
The XML-based IR represents a structured abstract syntax tree (AST) that explicitly encodes tool invocation order and inter-call dependencies:

\begin{astbox}
\begin{lstlisting}[language=xml,label={lst:xml_ast}, basicstyle=\ttfamily\small, breaklines=true]
<api id="0">
    <name>API_NAME_0</name>
    <param name="P_0" type="string">VALUE_0</param>
    <param name="P_1" type="integer">VALUE_1</param>
</api>

<api id="1">
    <name>API_NAME_1</name>
    <param name="P_0">
        <response api_id="0" api_name="API_NAME_0" type="placeholder"/>
    </param>
</api>
\end{lstlisting}
\end{astbox}

The execution order is encoded through explicit API identifiers, while inter-call dependencies are captured via response references, enabling deterministic execution of composed tool calls.

\subsection{JSON-Based AST Format.}
We additionally support a JSON-based AST format, where tool calls are represented as an ordered mapping from call identifiers to function invocations:

\begin{astbox}
\begin{lstlisting}[basicstyle=\ttfamily\small]
{
  "0": {
    "API_NAME_0": {
      "PARAM_NAME_0": VALUE_0,
      "PARAM_NAME_1": VALUE_1
    }
  },
  "1": {
    "API_NAME_1": {
      "PARAM_NAME_0": "API_RESPONSE_0"
    }
  }
}
\end{lstlisting}
\end{astbox}

In this format, dependencies between calls are expressed by referencing prior API responses. The parser validates JSON well-formedness, function names, parameter schemas, and data types before execution.

\subsection{Direct Call List Format.}
For simpler tool-use scenarios, we also support a direct call format, where tool invocations are expressed as a comma-separated list within brackets:

\begin{astbox}
\begin{lstlisting}[language=python, basicstyle=\ttfamily\small]
[
  API_NAME_0(PARAM_NAME_0=VALUE_0,
             PARAM_NAME_1=VALUE_1),
  API_NAME_1(PARAM_NAME_0=API_RESPONSE_0)
]
\end{lstlisting}
\end{astbox}

This format provides a lightweight representation for sequential tool calls without explicit AST structure. The parser infers execution order from list position and resolves dependencies by matching referenced outputs.

\section{Function Composition Examples}
\label{appendix:composition_structures}
We illustrate function composition of varying depths using representative examples. Here, the number of steps refers to the dependency depth of tool calls.

\subsection{Linear chaining}
Here, all calls are sequentially dependent, forming a chain \( A \rightarrow B \rightarrow C \).

\begin{promptbox}
Get all car listings from nearby dealerships.
\end{promptbox}
\begin{astbox}
\begin{lstlisting}[
    basicstyle=\ttfamily\small]
{
  "0": {
    "Get_current_location": {}
  },
  "1": {
    "Find_Dealerships_by_Location": {
      "location": "API_RESPONSE_0"
    }
  },
  "2": {
    "Get_Car_Listing_by_Dealerships": {
      "dealerships": "API_RESPONSE_1"
    }
  }
}
\end{lstlisting}
\end{astbox}

\subsection{Funnel Structure}
In this setting, intermediate calls are independent and jointly inform the final call, following the structure \( (A, B) \rightarrow C \).
\begin{promptbox}
    What is the current weather?
\end{promptbox}
\begin{astbox}
\begin{lstlisting}[basicstyle=\ttfamily\small]
{
  "0": {
    "Get_current_time": {}
  },
  "1": {
    "Get_current_location": {}
  },
  "2": {
    "Get_weather": {
      "time": "API_RESPONSE_0",
      "location": "API_RESPONSE_1"
    }
  }
}
\end{lstlisting}
\end{astbox}

\subsection{Tree Structure}
This structure requires a single call that is further branched out into multiple parallel calls \(A \rightarrow (B, C)\). 

\begin{promptbox}
    What is the price of bitcoin on Jan, 11, 2026. Give me its value in British pounds, and in Euros?
\end{promptbox}
\begin{astbox}
\begin{lstlisting}[basicstyle=\ttfamily\small]
{
  "0": {
    "Get_Crypto_Price": {
        "ticker": "BTC",
        "price_time": "2026-01-11"
    }
  },
  "1": {
    "Currency_conversion_API": {
        "amount": "API_RESPONSE_0",
        "from_currency": "USD",
        "to_currency": "GBP",
        "price_time": "2026-01-11"
    }
  },
  "2": {
    "Currency_conversion_API": {
        "amount": "API_RESPONSE_0",
        "from_currency": "USD",
        "to_currency": "EUR",
        "price_time": "2026-01-11"
    }
  }
}
\end{lstlisting}
\end{astbox}

\subsection{Conjunctive Constraint Composition}
This scenario is a constraint-based composition over a shared entity, where multiple tool calls impose independent predicates on the same object. 

\begin{promptbox}
    Show me Japanese restaurants in San Diego that have ratings above 4.2.
\end{promptbox}

\textbf{Option A:}
\begin{astbox}
\begin{lstlisting}[basicstyle=\ttfamily\small]

{
  "0": {
    "Find_Restaurants_by_Location": {
        "location": "San Diego"
    }
  },
  "1": {
    "Filter_Restaurants_by_Cuisine": {
        "restaurants": "API_RESPONSE_0",
        "cuisine": "Japanese"
    }
  },
  "2": {
    "Filter_Restaurants_by_ratings": {
        "restaurants": "API_RESPONSE_1",
        "ratings": 4.2
    }
  }
}
\end{lstlisting}
\end{astbox}

\textbf{Option B:}
\begin{astbox}
\begin{lstlisting}[basicstyle=\ttfamily\small]
{
  "0": {
    "Find_Restaurants_by_Location": {
        "location": "San Diego"
    }
  },
  "1": {
    "Filter_Restaurants_by_ratings": {
        "restaurants": "API_RESPONSE_0",
        "ratings": 4.2
    }
  },
  "2": {
    "Filter_Restaurants_by_Cuisine": {
        "restaurants": "API_RESPONSE_1",
        "cuisine": "Japanese"
    }
  }
}
\end{lstlisting}
\end{astbox}

Both syntaxes are valid since they could be applied in any order without changing the final result
\subsection{Mixed structure}
We could further combine multiple structures into one question that requires complex function composition

\begin{promptbox}
Could you find Mexican restaurants near me that are open right now?
\end{promptbox}
\begin{astbox}
\begin{lstlisting}[basicstyle=\ttfamily\small]
{
  "0": {
    "Get_current_location": { }
  },
  "1": {
    "Find_Restaurants_by_Location": {
        "location": "API_RESPONSE_0"
    }
  },
  "2": {
    "Filter_Restaurants_by_Cuisine": {
        "restaurants": "API_RESPONSE_1",
        "cuisine": "Mexican"
    }
  },
  "3": {
    "Get_current_time": { }
  }, 
  "4": {
    "Filter_Restaurants_by_Opening_Hours": {
        "restaurants": "API_RESPONSE_2",
        "time": "API_RESPONSE_3"
    }
  }
}
\end{lstlisting}  
\end{astbox}

\section{System Prompt}

We show an example system prompt used for JSON AST format

\begin{syspromptbox}
\begin{lstlisting}[basicstyle=\ttfamily\small,breaklines=true,
breakatwhitespace=true,
columns=fullflexible,
keepspaces=true]

# Instruction

You are an intelligent agent capable of advanced tool use. You observe your environment, reason deeply about it, and use tools or APIs at your disposal to answer user queries. Queries may require single or multiple API calls. Multiple calls may be executed in parallel (returning all outputs) or chained sequentially (returning only the final output).

## API Call Scenarios

1. **Single API call**: The query requires execution of a single API whose output will be returned to the user. (one output)

2. **Multiple Parallel API calls**: The query requires execution of a set of API calls simultaneously, and the outputs of all will be returned to the user. (all outputs)

3. **Multiple Chained API calls**: The query requires APIs to be chained together, where the output of one API serves as the input to another, and only the output of the final API call in the chain is returned to the user. (one output)

4. **Multi-step API calls (Single Query)**: A single query requires multiple sequential API calls. You respond with ONE API call at a time and wait for the environment to execute it and provide you with the tool response. If the tool response indicates that the problem is solved, return a NO-CALL response. If the tool call fails, retry it. If the tool call succeeds, proceed with the next call in the sequence.

5. **Multi-turn Conversations**: An extended conversation where the user provides multiple queries across several turns. Each turn may require multi-step API calls (as described in scenario 4). You must retain and utilize contextual information from previous exchanges to handle follow-up queries effectively. After completing one user's request with a NO-CALL response, the user may follow up with related queries or entirely new tasks. Use information from previous turns when relevant to the current query.

## Your Objectives

1. Determine whether the query can be solved using the APIs available to you. If not, inform the user accordingly with a NO-CALL response.
2. If the query can be solved, determine whether it requires a single API call, multiple parallel calls, multiple chained calls, or multi-step sequential calls.
3. Identify the appropriate API(s) from the available list and prepare a plan for solving the problem.
4. Determine the correct order of execution when chaining or sequencing is required.
5. Construct the final solution using the exact JSON format provided below.

## Response Format

You must strictly respond in one of the following JSON formats:

### Format 1: Single API Call

<think>
Explain your thought process to solve the question.
</think>
<tool_call return="one">
    {
        "0": {
            "API_NAME": {
                "PARAM_NAME_0": PARAM_VALUE_0,
                "PARAM_NAME_1": PARAM_VALUE_1
            }
        }
    }
</tool_call>

### Format 2: Multiple Parallel API Calls

<think>
Explain your thought process to solve the question.
</think>
<tool_call return="all">
    {
        "0": {
            "API_NAME_0": {
                "PARAM_NAME_0": PARAM_VALUE_0
            }
        },
        "1": {
            "API_NAME_1": {
                "PARAM_NAME_0": PARAM_NAME_0
            }
        }
    }
</tool_call>

### Format 3: Multiple Chained API Calls

<think>
Explain your thought process to solve the question.
</think>
<tool_call return="one">
    {
        "0": {
            "API_NAME_0": {
                "PARAM_NAME_0": PARAM_VALUE_0
            }
        },
        "1": {
            "API_NAME_1": {
                "PARAM_NAME_0": PARAM_NAME_0,
                "PARAM_NAME_1": "API_RESPONSE_0"
            }
        }
    }
</tool_call>

### Format 4: Multi-step API Calls (One at a Time)
For multi-step scenarios, respond with ONLY ONE API call per response:

<think>
Explain the current step in your plan.
</think>
<tool_call return="one">
    {
        "0": {
            "API_NAME": {
                "PARAM_NAME_0": PARAM_VALUE_0
            }
        }
    }
</tool_call>

After receiving the tool response, proceed with the next step or return NO-CALL if complete.

### Format 5: NO-CALL Response
Use this format when:
- The query cannot be solved using available APIs
- The problem has already been solved based on previous tool responses
- No further action is required

<think>
Explain why no API calls are required. Specify either: (1) the query cannot be solved using the available APIs, or (2) the problem has been solved based on the environment response.
</think>
<tool_call return="one">
    {}
</tool_call>

# Rules

1. **API Usage**: Use only the provided APIs. Do not invent, assume, or reference APIs that are not explicitly listed.

2. **Return Attribute**: Indicate whether the output of all APIs needs to be returned or only one output needs to be returned using the "return" attribute of the `<tool_call>` tag:
   - `<tool_call return="all">`: All outputs will be returned to the user. Use this for parallel calls.
   - `<tool_call return="one">`: Only the output of the final API call will be returned to the user. Use this for single, chained, or multi-step calls.

3. **Parameters**: Include all required parameters in each API call. Optional parameters may be added if they improve the result or are necessary to fulfill the query.

4. **API IDs**: Each API call must have a unique numeric string key. The keys should be zero-indexed (starting from "0").

5. **ID Sequencing**: The keys used should be a sequence of increasing integers like "0", "1", "2", "3", etc. For chained calls, this sequence represents the order of execution.

6. **JSON Structure**: The tool call JSON must be a dictionary where each key is a numeric string ID and each value is an object containing a single API name as key and its parameters object as value.

7. **API Name**: Each API call object (e.g., `{"0": {...}}`) must contain exactly one API name as the key, with its parameters as the value.

8. **Parameter Format**: Parameters for each API should be provided as a JSON object with parameter names as keys and their values as the corresponding values.

9. **Parameter Types**: API parameter values may be primitive types (boolean, integer, float, string), complex types (dict, array), or string placeholders referencing previous API responses.

10. **Response References**: To reference the output of a previous API call in chained scenarios, use the string value `"API_RESPONSE_X"` where X is the numeric ID of the API whose output you want to reference.

11. **Chaining APIs**: When chaining API calls together, use `"API_RESPONSE_0"`, `"API_RESPONSE_1"`, etc. as parameter values to pass the output from one API as input to another.

12. **Multi-step Execution**: For multi-step scenarios (scenarios 4 and 5), respond with only ONE API call at a time (a JSON object with a single key "0"). Wait for the environment to execute it and provide feedback before proceeding with the next step.

13. **Context Retention**: For multi-turn conversations (scenario 5), maintain awareness of previous exchanges and use that context when relevant to the current query.

14. **NO-CALL Format**: When no API calls are needed, respond with an empty JSON object `{}` inside the `<tool_call return="one">` tag.

\end{lstlisting}
\end{syspromptbox}

The available tool definitions are added at the end of this prompt dynamically. In addition to the required tools we also add irrelevant and ambiguous tool definitions to elicit reasoning capabilities of the model.

\section{Traces from the trained model}

\begin{promptbox}
Make reservation at HTL\_LONDON\_PARK for 7 nights from August 1 to August 8, 2024 for 3 people. Guest Robert Wilson, phone 555-0567, email rob@vacation.com. Charge credit card 3782-822463-10005, billing address 789 Pine St, Los Angeles for Robert Wilson.
\end{promptbox}

\begin{modelbox}
\begin{lstlisting}[
  basicstyle=\ttfamily\small,
  breaklines=true,
  breakatwhitespace=true,
  columns=fullflexible,
  keepspaces=true
]
<think>
Okay, let's tackle this user request. The user wants to make a hotel reservation at HTL_LONDON_PARK for 7 nights from August 1 to August 8, 2024, for 3 people. The guest details are Robert Wilson, phone 555-0567, email rob@vacation.com. They also want to charge a credit card: 3782-822463-10005, with billing address 789 Pine St, Los Angeles.
First, I need to check which API to use. The available APIs are GetDate, BookHotel, Doordash.place_order, and BeanAndBrewCoffeehouse.get_opening_hours. The task is about booking a hotel, so the BookHotel API is the right choice.
Looking at the parameters required for BookHotel: hotel_id, number_of_nights, check_details (with check_in and check_out dates), number_of_people, personal_details (name, phone, email), and billing_details (name, payment_method, and credit info if applicable).

The user provided all the necessary details. The check-in date is August 1, 2024, and check-out is August 8, 2024. The number of nights is 7, which matches the date difference. The hotel_id is HTL_LONDON_PARK. Number of people is 3. Personal details include Robert Wilson, phone 555-0567, and email rob@vacation.com. Billing details require the credit card info and address. The payment method is credit, so the credit parameters need to be included.

Wait, the credit card number is given as 3782-822463-10005. The API might require it without hyphens, but the user provided it with hyphens. However, the parameter type is a string, so including the hyphens should be okay. The billing address is 789 Pine St, Los Angeles, and the billing name is Robert Wilson.

I need to structure the parameters correctly. The check_details should have check_in as "2024-08-01" and check_out as "2024-08-08". The number_of_nights is 7. The personal_details name is Robert Wilson, phone_number is 555-0567, and email is rob@vacation.com. The billing_details name is Robert Wilson, payment_method is "credit", and the credit object includes address, card_number, and city. The city is Los Angeles.

Wait, the billing address is 789 Pine St, Los Angeles. So the address field in the credit object should be "789 Pine St, Los Angeles", and the city is "Los Angeles". That makes sense. The card number is 3782-822463-10005, which I'll input as-is since it's a string.

Now, verifying all required parameters are present. The hotel_id is provided, number_of_nights is 7, check_details with both dates, number_of_people is 3, personal_details with name and phone, and billing_details with all required fields. All required parameters are covered.

So, the BookHotel API call should include all these parameters. There's no need for other APIs here. The user isn't asking for dates, ordering food, or coffeehouse hours. So this single API call should handle the reservation.
</think>

<tool_call return="one">
{
  "0": {
    "BookHotel": {
      "hotel_id": "HTL_LONDON_PARK",
      "number_of_nights": 7,
      "check_details": {
        "check_in": "2024-08-01",
        "check_out": "2024-08-08"
      },
      "number_of_people": 3,
      "personal_details": {
        "name": "Robert Wilson",
        "phone_number": "555-0567",
        "email": "rob@vacation.com"
      },
      "billing_details": {
        "name": "Robert Wilson",
        "payment_method": "credit",
        "credit": {
          "address": "789 Pine St, Los Angeles",
          "card_number": "3782-822463-10005",
          "city": "Los Angeles"
        }
      }
    }
  }
}
</tool_call>
\end{lstlisting}
\end{modelbox}

\begin{promptbox}
Can you find open French restaurants in my vicinity?
\end{promptbox}

\begin{modelbox}
\begin{lstlisting}[
  basicstyle=\ttfamily\small,
  breaklines=true,
  breakatwhitespace=true,
  columns=fullflexible,
  keepspaces=true
]
<think>
Okay, let's see. The user is asking for open French restaurants in their vicinity. First, I need to figure out what APIs to use here.

Hmm, the available APIs include GetCurrentLocation, which can get the user's location. Then there's FindRestaurantsByLocation, which uses a location parameter to find restaurants. But the user wants French cuisine, so maybe after getting the restaurants, I need to filter by cuisine using FilterByCuisine. Also, they want open restaurants, so I might need to check the opening hours with FilterByOpeningHours. But wait, the current time is needed for that. So first, get the current time with GetTime. Then, get the location, then find restaurants in that location, filter by French cuisine, then check which are open now. But the user might not have provided a specific time, so maybe the opening hours are checked against the current time. Let me outline the steps:

1. Get current time using GetTime.
2. Get current location using GetCurrentLocation.
3. Use FindRestaurantsByLocation with the location from step 2 to get all restaurants there.
4. Filter those restaurants to French cuisine using FilterByCuisine.
5. Then, filter the French restaurants by opening hours using FilterByOpeningHours, using the current time from step 1.

Wait, but the FilterByOpeningHours requires a specific time. So after getting the current time, I can pass that as the time parameter. So the steps are sequential. Each step depends on the previous one's output. So this would be a chained API call scenario. Let me check the parameters. 
For example, FindRestaurantsByLocation needs a location, which comes from GetCurrentLocation. Then FilterByCuisine needs the cuisine (French) and the restaurants DataFrame from FindRestaurantsByLocation. Then FilterByOpeningHours needs the current time and the filtered restaurants from FilterByCuisine. So the order is important. So the first call is GetCurrentLocation, then FindRestaurantsByLocation with that location, then FilterByCuisine with cuisine=French and the restaurants from the previous step, then FilterByOpeningHours with time from GetTime and the restaurants from the previous filter. 
Then, the final result would be the open French restaurants. But the user might not have specified a time, so the opening hours check is necessary. However, if the current time is not provided, the FilterByOpeningHours can't be done. Therefore, the steps need to be in that order. So the first API call should be GetCurrentLocation to get the location. Then FindRestaurantsByLocation with that location. Then FilterByCuisine for French. Then FilterByOpeningHours with the current time. But the current time needs to be obtained first. Wait, the GetTime API can be called first to get the current time. Then GetCurrentLocation. Then FindRestaurantsByLocation. Then FilterByCuisine. Then FilterByOpeningHours using the time from GetTime. So the order is: GetTime, GetCurrentLocation, FindRestaurantsByLocation, FilterByCuisine, FilterByOpeningHours. But the user's query is to find open French restaurants, so the final step would be FilterByOpeningHours. 
However, the user might not have a specific time, so the current time is used. Therefore, the correct sequence is to first get the current time, then the location, then find restaurants in that location, filter by French cuisine, then filter those by opening hours using the current time. So the API calls would be chained in that order. Let me check the parameters again. For example, the FilterByOpeningHours requires the time and the restaurants DataFrame. So after getting the French restaurants, we pass them to FilterByOpeningHours with the current time. So the steps are sequential. Therefore, the first API call should be GetTime to get the current time. Then GetCurrentLocation. Then FindRestaurantsByLocation with the location. Then FilterByCuisine with cuisine=French. Then FilterByOpeningHours with the time from GetTime and the restaurants from the previous step. So the final answer would be the filtered restaurants that are open now. But the user's query is to find open French restaurants, so the last step is necessary. Therefore, the correct approach is to chain these APIs in the order mentioned. So the first call is GetTime, then GetCurrentLocation, then FindRestaurantsByLocation, then FilterByCuisine, then FilterByOpeningHours. Each step's output is used as input for the next. Therefore, the tool calls would be in sequence. Let me structure this as a multi-step API call scenario. The first step is to get the current time. Then the location. Then find restaurants by location. Then filter by cuisine. Then filter by opening hours. So the initial call would be GetTime, then GetCurrentLocation, and so on. But the user might not have specified a time, so the current time is necessary. 
Therefore, the first API call is GetTime. Then, the next is GetCurrentLocation. Then, using the location from GetCurrentLocation, call FindRestaurantsByLocation. Then, filter those with FilterByCuisine for French. Then, use the current time from GetTime to filter by opening hours. So the final answer would be the list of open French restaurants. Therefore, the correct sequence of API calls is GetTime, GetCurrentLocation, FindRestaurantsByLocation, FilterByCuisine, FilterByOpeningHours. Each step depends on the previous one's output. So the first tool call is GetTime.
</think>

<tool_call return="one">
    {
        "0": {
            "GetTime": {}
        },
        "1": {
            "GetCurrentLocation": {}
        },
        "2": {
            "FindRestaurantsByLocation": {
                "location": "API_RESPONSE_1"
            }
        },
        "3": {
            "FilterByCuisine": {
                "cuisine": "French",
                "restaurants": "API_RESPONSE_2"
            }
        },
        "4": {
            "FilterByOpeningHours": {
                "time": "API_RESPONSE_0",
                "restaurants": "API_RESPONSE_3"
            }
        }
    }
</tool_call>
\end{lstlisting}
\end{modelbox}

\section{Dataset Comparison and Analysis}
\label{appendix:dataset_analysis}

\subsection{Adapting Existing Tool-Use Datasets for TIER}
DepthBench is deliberately compact: 1.7K examples over 163 tools, compared to $\sim$10K examples over $\sim$26K tools in ToolACE and $\sim$60K examples over $\sim$4K tools in xLAM. To assess whether this small, composition-focused corpus is competitive with established large-scale tool-use datasets, we train Qwen3-8B under TIER on adapted versions of both, using the same reward formulation, optimizer, and rollout budget as in 
\Cref{sec:experiment_results}.

\subsection{Adaptation procedure}
ToolACE and xLAM are released as supervised fine-tuning corpora providing a single annotated trajectory per query, without an executable backend; neither natively supports execution-grounded rewards. We construct executable variants by parsing each function schema to recover its parameter set, declared types, and a required/optional partition. Because the two corpora encode this partition differently, we apply dataset-specific rules. ToolACE exposes an explicit required field alongside per-parameter defaults; we mark a parameter as required iff it appears in required, with the field taking precedence when a parameter both appears there and declares a default. xLAM exposes no required field, so we treat any parameter with a declared default as optional and all others as required.

A call passes parsing when every supplied argument belongs to the schema, all values type-check against their declarations, and every required parameter is present; optional parameters may be omitted. Parsed calls return a deterministic response that echoes the resolved arguments, with omitted optionals materialised to their declared defaults prior to echoing. Because the source corpora contain no golden tool outputs, this design isolates correctness at the parameter-passing layer; an omitted optional and the same optional passed at its default value yield identical responses by construction, so optional-parameter choices cannot influence downstream correctness through the execution layer. 

We score a call as correct under a relaxed match against the released trajectory: required parameters must match the gold values exactly, while optional parameters are not scored. The criterion is strictly stronger than parsing success and strictly weaker than full trajectory string match — the latter would penalise superficial reordering or default-value rewrites that leave behaviour unchanged.

\subsection{Scope}
DepthBench additionally contains multi-turn dialogues in which composition spans across turns; the present work focuses on single-turn composition, and the multi-turn split is excluded from both training and evaluation.
ToolACE additionally contains a multi-turn split. We restrict this work to single-turn composition and exclude multi-turn data from both training and evaluation; the same restriction applies to DepthBench's multi-turn split.
\subsection{Results}
\Cref{tab:dataset_comparison} reports BFCL v3 performance for models trained on each corpus under identical TIER supervision. Despite being roughly an order of magnitude smaller, the DepthBench-trained model achieves the strongest overall performance. The gap is concentrated on the Multi-Turn metric (39.1 vs. 24.1 and 21.2); single-turn metrics (Non-Live AST, Live) lie within a few points across the three corpora and do not separate them.
\subsection{Analysis}
Two dataset-level properties appear to drive the gap. First, xLAM contains frequent mismatches between released trajectories and the corresponding tool schemas — argument names, types, or values in the annotated trajectory do not match the declared signature. Under execution-grounded rewards, these inconsistencies penalize schema-conformant calls, because the schema and the trajectory disagree on what counts as correct. Second, ToolACE pairs a very large tool vocabulary ($\sim$26K) with relatively few examples ($\sim$10K), so most tools appear in at most one query. The resulting models invoke individual tools too rigidly to handle the argument variation present at evaluation time. Neither property is visible from single-call accuracy, which is why aggregate function-calling benchmarks fail to expose the gap that BFCL Multi-Turn surfaces.
\begin{table*}[t]
  \caption{Comparison between sizes of DepthBench and ToolAce}
  \label{tab:dataset_comparison}
  \begin{center}
    \begin{small}
      \begin{sc}
      \resizebox{\textwidth}{!}{
        \begin{tabular}{lcccccc}
          \toprule
          Dataset (size) & Overall & Non-Live AST        & Live   &  Multiturn & Relevance  & Irrelevance   \\
          \midrule
          
          DepthBench (1240) & \textbf{68.92} & 85.94 & 81.96 & \textbf{39.12} & \textbf{77.78} & 87.65 \\
          Toolace (10K) & 64.66 & 85.90 & \textbf{82.81} & 24.12 & 61.11 & \textbf{91.01} \\
          XLAM (60K) & 63.02 & \textbf{89.29} & 79.16 & 21.22 & \textbf{77.78} & 83.92 \\
          \bottomrule
        \end{tabular}
        }
      \end{sc}
    \end{small}
  \end{center}
  \vskip -0.1in
\end{table*}

\section{RL Algorithm Ablation}
\label{appendix:algorithm_ablation}

\begin{figure}[H]
  \centering
  \includegraphics[width=0.6 \columnwidth]{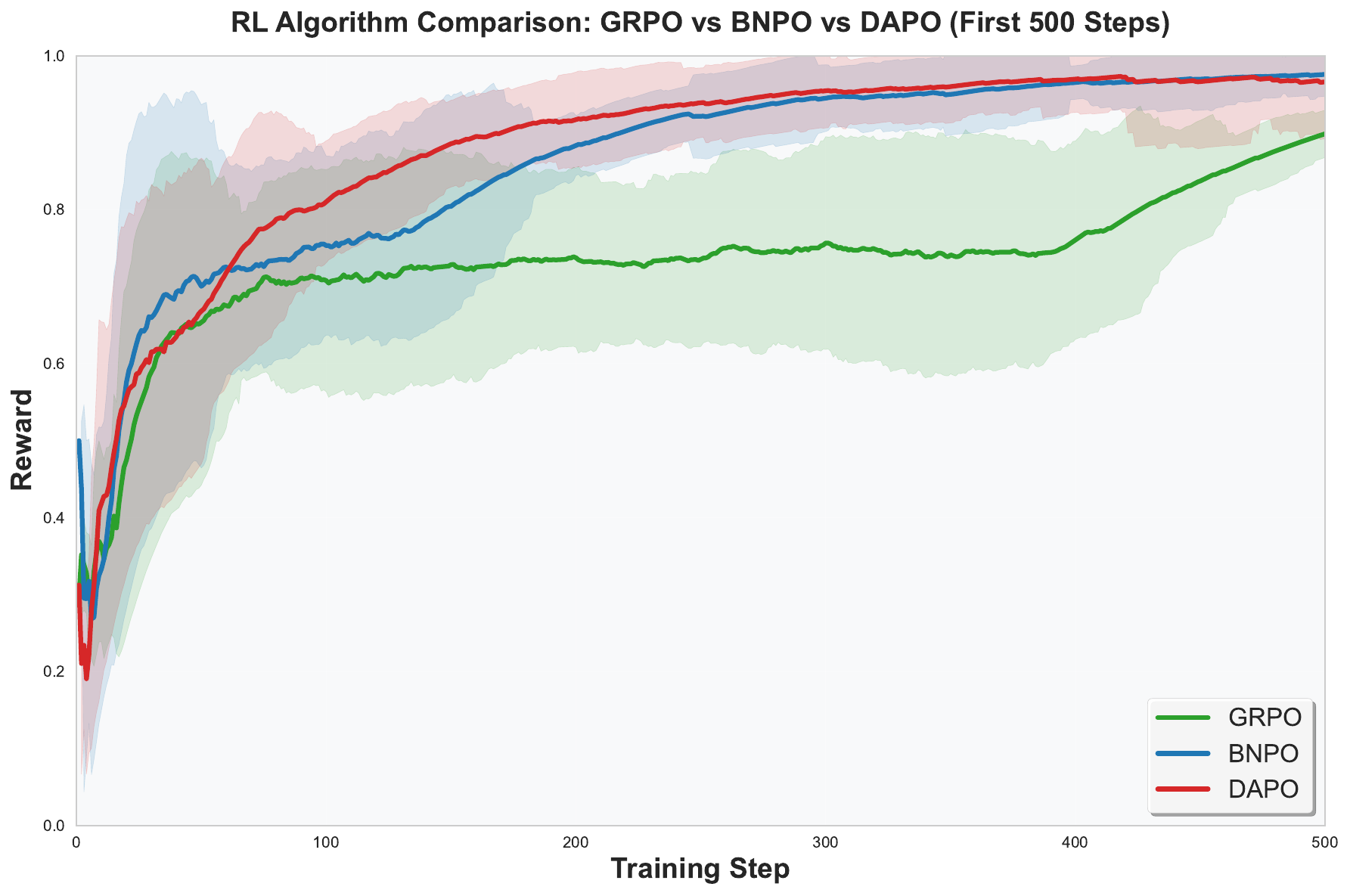}
  \caption{Training reward over the first 500 steps for three RL algorithms (Standard GRPO, Batch-Normalized GRPO, DAPO) under identical fine-grained, trajectory-invariant reward formulation. Shaded regions show $\pm1$ standard deviation. DAPO converges fastest, Batch-Normalized GRPO converges steadily to the highest level, and GRPO exhibits a delayed plateau-then-jump pattern. All three ultimately reach high reward, confirming that the reward design drives performance rather than the optimization algorithm.}
  \label{fig:algo_comparison}
\end{figure}

To verify that the gains reported in this paper are driven by reward design rather than the choice of optimization algorithm, we train three models using different policy optimization methods—Standard GRPO, Batch-Normalized GRPO, and DAPO—under identical reward formulation, base model (Qwen3-8B), rollout budget, and hyperparameters. The only variable across conditions is the optimization objective. \Cref{fig:algo_comparison} shows training reward curves as exponentially-weighted moving averages of per-step batch-mean reward logged over a 50-step rolling window (smoothing constant, $\alpha=0.01$). The shaded region is a $\pm1$ rolling standard deviation.

\subsection{Training Dynamics}

DAPO converges fastest, reaching approximately 0.9 reward by step~100 and stabilizing with moderate variance. Batch-Normalized GRPO converges more gradually, reaching near-ceiling reward by step~200 and continuing to improve through step~500. Standard GRPO exhibits a distinct trajectory: it plateaus around 0.72–0.75 for an extended period (steps 100–350), followed by a delayed but sharp improvement after step~400, eventually approaching 0.9. This plateau-then-jump behavior suggests that GRPO requires a longer exploration phase before discovering effective compositional strategies, likely due to its reliance on group-relative comparisons. GRPO also shows higher variance throughout training, indicating less stable optimization compared to Batch-Normalized GRPO and DAPO.

Despite these differences in convergence behavior, all three algorithms ultimately reach high reward levels, demonstrating that the fine-grained reward signal is learnable across optimization objectives.

\subsection{Evaluation Results}

We report accuracy on the composition benchmark stratified by depth in \Cref{tab:algo_ablation}.

\begin{table*}[t]
  \centering
  \caption{RL algorithm ablation on the composition benchmark. All configurations use fine-grained TIER rewards, base model, and rollout budget. Only the optimization objective differs.}
  \label{tab:algo_ablation}
  \small
  \setlength{\tabcolsep}{5pt}
  \resizebox{\textwidth}{!}{
    \begin{tabular}{lcccccccc}
      \toprule
      Algorithm & Overall Acc & 0-Step & 1-Step & 2-Step & 3-Step & 4-Step & 5-Step & 6-Step \\
      \midrule
      Standard GRPO & 97.17 & 98.04 & 98.10 & 95.00 & 96.47 & 98.93 & \textbf{97.78} & \textbf{95.26} \\
      Batch-Normalized GRPO & \textbf{98.56} & 98.04 & \textbf{99.38} & 97.89 & \textbf{100.00} & 93.33 & 90.00 & 90.00 \\
      DAPO & 93.21 & \textbf{100.00} & 94.79 & \textbf{98.95} & 76.00 & 88.89 & \textbf{100.00} & 70.00 \\
      \bottomrule
    \end{tabular}
  }
\end{table*}

All three algorithms achieve over 90\% overall accuracy, confirming that the fine-grained TIER reward design is the primary driver of compositional capability rather than any specific optimization method. 

However, the algorithms differ in their depth profiles. Batch-Normalized GRPO achieves the highest overall accuracy (98.56\%) and the most consistent performance across depths (90–100\%). Standard GRPO achieves comparable overall performance (97.17\%) with particularly strong results at deeper compositions (97.78\% at 5-step and 95.26\% at 6-step), despite slower convergence during training. DAPO exhibits the greatest variance across depths: while it achieves perfect accuracy at 0-step and 100\% at 5-step, it drops to 76.00\% at 3-step and 70.00\% at 6-step.

This variance in DAPO aligns with its faster but less stable convergence—rapid policy commitment yields strong performance on some compositions but weaker coverage of others. In contrast, GRPO’s extended exploration phase appears to improve coverage across compositional patterns, leading to more uniform depth performance.

Based on these results, we adopt Batch-Normalized GRPO as the default optimization method due to its strong overall performance and consistency across depths. The key finding, however, is that \emph{all three algorithms enable multi-step composition when paired with fine-grained, trajectory-invariant, execution rewards}. A reward design that yields near-zero accuracy at 5–6 steps under trajectory-supervised training instead enables high multi-step performance across all optimization methods, with GRPO and Batch-Normalized GRPO achieving consistently high accuracy and DAPO showing more variable but still substantial gains.

\section{Training Compute}
\label{appendix:compute}
All training runs were conducted on an internal cluster using 2× NVIDIA A100 (80GB) GPUs per run. Each run trained for 1,000 optimization steps and required approximately 20 wall-clock hours ($\approx$40 GPU-hours per run). In total, we trained well over 100 model configurations, corresponding to roughly 4,000 GPU-hours of training compute. Each trained model was subsequently evaluated on BFCL and DepthBench using a single A100 (80GB) GPU, amounting to approximately 1,680 GPU-hours of evaluation compute. The full research project additionally incurred approximately 1,500 USD in OpenAI API costs across dataset evaluation, exploratory evaluations on ComplexFuncBench that we ultimately excluded from the main results (see \Cref{sec:limitations}). Beyond the runs reported above, additional GPU compute was spent on preliminary and discarded experiments that are not included in this paper.

%% file: references.bib
@misc{qwen3,
      title={Qwen3 Technical Report}, 
      author={An Yang and Anfeng Li and Baosong Yang and Beichen Zhang and Binyuan Hui and Bo Zheng and Bowen Yu and Chang Gao and Chengen Huang and Chenxu Lv and Chujie Zheng and Dayiheng Liu and Fan Zhou and Fei Huang and Feng Hu and Hao Ge and Haoran Wei and Huan Lin and Jialong Tang and Jian Yang and Jianhong Tu and Jianwei Zhang and Jianxin Yang and Jiaxi Yang and Jing Zhou and Jingren Zhou and Junyang Lin and Kai Dang and Keqin Bao and Kexin Yang and Le Yu and Lianghao Deng and Mei Li and Mingfeng Xue and Mingze Li and Pei Zhang and Peng Wang and Qin Zhu and Rui Men and Ruize Gao and Shixuan Liu and Shuang Luo and Tianhao Li and Tianyi Tang and Wenbiao Yin and Xingzhang Ren and Xinyu Wang and Xinyu Zhang and Xuancheng Ren and Yang Fan and Yang Su and Yichang Zhang and Yinger Zhang and Yu Wan and Yuqiong Liu and Zekun Wang and Zeyu Cui and Zhenru Zhang and Zhipeng Zhou and Zihan Qiu},
      year={2025},
      eprint={2505.09388},
      archivePrefix={arXiv},
      primaryClass={cs.CL},
      url={https://arxiv.org/abs/2505.09388}, 
}

@misc{yao23,
      title={ReAct: Synergizing Reasoning and Acting in Language Models}, 
      author={Shunyu Yao and Jeffrey Zhao and Dian Yu and Nan Du and Izhak Shafran and Karthik Narasimhan and Yuan Cao},
      year={2023},
      eprint={2210.03629},
      archivePrefix={arXiv},
      primaryClass={cs.CL},
}

@article{wei23,
  title={Chain-of-thought prompting elicits reasoning in large language models},
  author={Wei, Jason and Wang, Xuezhi and Schuurmans, Dale and Bosma, Maarten and Xia, Fei and Chi, Ed and Le, Quoc V and Zhou, Denny and others},
  journal={Advances in neural information processing systems},
  volume={35},
  pages={24824--24837},
  year={2022}
}

@misc{xlam,
      title={xLAM: A Family of Large Action Models to Empower AI Agent Systems}, 
      author={Jianguo Zhang and Tian Lan and Ming Zhu and Zuxin Liu and Thai Hoang and Shirley Kokane and Weiran Yao and Juntao Tan and Akshara Prabhakar and Haolin Chen and Zhiwei Liu and Yihao Feng and Tulika Awalgaonkar and Rithesh Murthy and Eric Hu and Zeyuan Chen and Ran Xu and Juan Carlos Niebles and Shelby Heinecke and Huan Wang and Silvio Savarese and Caiming Xiong},
      year={2024},
      eprint={2409.03215},
      archivePrefix={arXiv},
      primaryClass={cs.CL},
}

@misc{toolace,
      title={ToolACE: Winning the Points of LLM Function Calling}, 
      author={Weiwen Liu and Xu Huang and Xingshan Zeng and Xinlong Hao and Shuai Yu and Dexun Li and Shuai Wang and Weinan Gan and Zhengying Liu and Yuanqing Yu and Zezhong Wang and Yuxian Wang and Wu Ning and Yutai Hou and Bin Wang and Chuhan Wu and Xinzhi Wang and Yong Liu and Yasheng Wang and Duyu Tang and Dandan Tu and Lifeng Shang and Xin Jiang and Ruiming Tang and Defu Lian and Qun Liu and Enhong Chen},
      year={2025},
      eprint={2409.00920},
      archivePrefix={arXiv},
      primaryClass={cs.LG},
}

@article{Guo25,
   title={DeepSeek-R1 incentivizes reasoning in LLMs through reinforcement learning},
   volume={645},
   ISSN={1476-4687},
   DOI={10.1038/s41586-025-09422-z},
   number={8081},
   journal={Nature},
   publisher={Springer Science and Business Media LLC},
   author={Guo, Daya and Yang, Dejian and Zhang, Haowei and Song, Junxiao and Wang, Peiyi and Zhu, Qihao and Xu, Runxin and Zhang, Ruoyu and Ma, Shirong and Bi, Xiao and Zhang, Xiaokang and Yu, Xingkai and Wu, Yu and Wu, Z. F. and Gou, Zhibin and Shao, Zhihong and Li, Zhuoshu and Gao, Ziyi and Liu, Aixin and Xue, Bing and Wang, Bingxuan and Wu, Bochao and Feng, Bei and Lu, Chengda and Zhao, Chenggang and Deng, Chengqi and Ruan, Chong and Dai, Damai and Chen, Deli and Ji, Dongjie and Li, Erhang and Lin, Fangyun and Dai, Fucong and Luo, Fuli and Hao, Guangbo and Chen, Guanting and Li, Guowei and Zhang, H. and Xu, Hanwei and Ding, Honghui and Gao, Huazuo and Qu, Hui and Li, Hui and Guo, Jianzhong and Li, Jiashi and Chen, Jingchang and Yuan, Jingyang and Tu, Jinhao and Qiu, Junjie and Li, Junlong and Cai, J. L. and Ni, Jiaqi and Liang, Jian and Chen, Jin and Dong, Kai and Hu, Kai and You, Kaichao and Gao, Kaige and Guan, Kang and Huang, Kexin and Yu, Kuai and Wang, Lean and Zhang, Lecong and Zhao, Liang and Wang, Litong and Zhang, Liyue and Xu, Lei and Xia, Leyi and Zhang, Mingchuan and Zhang, Minghua and Tang, Minghui and Zhou, Mingxu and Li, Meng and Wang, Miaojun and Li, Mingming and Tian, Ning and Huang, Panpan and Zhang, Peng and Wang, Qiancheng and Chen, Qinyu and Du, Qiushi and Ge, Ruiqi and Zhang, Ruisong and Pan, Ruizhe and Wang, Runji and Chen, R. J. and Jin, R. L. and Chen, Ruyi and Lu, Shanghao and Zhou, Shangyan and Chen, Shanhuang and Ye, Shengfeng and Wang, Shiyu and Yu, Shuiping and Zhou, Shunfeng and Pan, Shuting and Li, S. S. and Zhou, Shuang and Wu, Shaoqing and Yun, Tao and Pei, Tian and Sun, Tianyu and Wang, T. and Zeng, Wangding and Liu, Wen and Liang, Wenfeng and Gao, Wenjun and Yu, Wenqin and Zhang, Wentao and Xiao, W. L. and An, Wei and Liu, Xiaodong and Wang, Xiaohan and Chen, Xiaokang and Nie, Xiaotao and Cheng, Xin and Liu, Xin and Xie, Xin and Liu, Xingchao and Yang, Xinyu and Li, Xinyuan and Su, Xuecheng and Lin, Xuheng and Li, X. Q. and Jin, Xiangyue and Shen, Xiaojin and Chen, Xiaosha and Sun, Xiaowen and Wang, Xiaoxiang and Song, Xinnan and Zhou, Xinyi and Wang, Xianzu and Shan, Xinxia and Li, Y. K. and Wang, Y. Q. and Wei, Y. X. and Zhang, Yang and Xu, Yanhong and Li, Yao and Zhao, Yao and Sun, Yaofeng and Wang, Yaohui and Yu, Yi and Zhang, Yichao and Shi, Yifan and Xiong, Yiliang and He, Ying and Piao, Yishi and Wang, Yisong and Tan, Yixuan and Ma, Yiyang and Liu, Yiyuan and Guo, Yongqiang and Ou, Yuan and Wang, Yuduan and Gong, Yue and Zou, Yuheng and He, Yujia and Xiong, Yunfan and Luo, Yuxiang and You, Yuxiang and Liu, Yuxuan and Zhou, Yuyang and Zhu, Y. X. and Huang, Yanping and Li, Yaohui and Zheng, Yi and Zhu, Yuchen and Ma, Yunxian and Tang, Ying and Zha, Yukun and Yan, Yuting and Ren, Z. Z. and Ren, Zehui and Sha, Zhangli and Fu, Zhe and Xu, Zhean and Xie, Zhenda and Zhang, Zhengyan and Hao, Zhewen and Ma, Zhicheng and Yan, Zhigang and Wu, Zhiyu and Gu, Zihui and Zhu, Zijia and Liu, Zijun and Li, Zilin and Xie, Ziwei and Song, Ziyang and Pan, Zizheng and Huang, Zhen and Xu, Zhipeng and Zhang, Zhongyu and Zhang, Zhen},
   year={2025},
   month=sep, pages={633–638} }

@misc{qian25,
      title={ToolRL: Reward is All Tool Learning Needs}, 
      author={Cheng Qian and Emre Can Acikgoz and Qi He and Hongru Wang and Xiusi Chen and Dilek Hakkani-Tür and Gokhan Tur and Heng Ji},
      year={2025},
      eprint={2504.13958},
      archivePrefix={arXiv},
      primaryClass={cs.LG},
}

@article{paranjape2023art,
  title={Art: Automatic multi-step reasoning and tool-use for large language models},
  author={Paranjape, Bhargavi and Lundberg, Scott and Singh, Sameer and Hajishirzi, Hannaneh and Zettlemoyer, Luke and Ribeiro, Marco Tulio},
  journal={arXiv preprint arXiv:2303.09014},
  year={2023}
}

@article{zhuang2023toolchain,
  title={Toolchain*: Efficient action space navigation in large language models with a* search},
  author={Zhuang, Yuchen and Chen, Xiang and Yu, Tong and Mitra, Saayan and Bursztyn, Victor and Rossi, Ryan A and Sarkhel, Somdeb and Zhang, Chao},
  journal={arXiv preprint arXiv:2310.13227},
  year={2023}
}

@article{song2023restgpt,
  title={Restgpt: Connecting large language models with real-world restful apis},
  author={Song, Yifan and Xiong, Weimin and Zhu, Dawei and Wu, Wenhao and Qian, Han and Song, Mingbo and Huang, Hailiang and Li, Cheng and Wang, Ke and Yao, Rong and others},
  journal={arXiv preprint arXiv:2306.06624},
  year={2023}
}

@article{schick2023toolformer,
  title={Toolformer: Language models can teach themselves to use tools},
  author={Schick, Timo and Dwivedi-Yu, Jane and Dess{\`\i}, Roberto and Raileanu, Roberta and Lomeli, Maria and Hambro, Eric and Zettlemoyer, Luke and Cancedda, Nicola and Scialom, Thomas},
  journal={Advances in Neural Information Processing Systems},
  volume={36},
  pages={68539--68551},
  year={2023}
}

@article{liang2024taskmatrix,
  title={Taskmatrix. ai: Completing tasks by connecting foundation models with millions of apis},
  author={Liang, Yaobo and Wu, Chenfei and Song, Ting and Wu, Wenshan and Xia, Yan and Liu, Yu and Ou, Yang and Lu, Shuai and Ji, Lei and Mao, Shaoguang and others},
  journal={Intelligent Computing},
  volume={3},
  pages={0063},
  year={2024},
  publisher={AAAS}
}

@article{liu2024apigen,
  title={Apigen: Automated pipeline for generating verifiable and diverse function-calling datasets},
  author={Liu, Zuxin and Hoang, Thai and Zhang, Jianguo and Zhu, Ming and Lan, Tian and Tan, Juntao and Yao, Weiran and Liu, Zhiwei and Feng, Yihao and RN, Rithesh and others},
  journal={Advances in Neural Information Processing Systems},
  volume={37},
  pages={54463--54482},
  year={2024}
}

@article{qin2023toolllm,
  title={Toolllm: Facilitating large language models to master 16000+ real-world apis},
  author={Qin, Yujia and Liang, Shihao and Ye, Yining and Zhu, Kunlun and Yan, Lan and Lu, Yaxi and Lin, Yankai and Cong, Xin and Tang, Xiangru and Qian, Bill and others},
  journal={arXiv preprint arXiv:2307.16789},
  year={2023}
}

@article{du2025generalizable,
  title={Generalizable end-to-end tool-use rl with synthetic codegym},
  author={Du, Weihua and Gong, Hailei and Ling, Zhan and Liu, Kang and Shen, Lingfeng and Yao, Xuesong and Xu, Yufei and Shi, Dingyuan and Yang, Yiming and Chen, Jiecao},
  journal={arXiv preprint arXiv:2509.17325},
  year={2025}
}

@inproceedings{qi2024learning,
  title={Learning generalizable tool-use skills through trajectory generation},
  author={Qi, Carl and Wu, Yilin and Yu, Lifan and Liu, Haoyue and Jiang, Bowen and Lin, Xingyu and Held, David},
  booktitle={2024 IEEE/RSJ International Conference on Intelligent Robots and Systems (IROS)},
  pages={2847--2854},
  year={2024},
  organization={IEEE}
}

@inproceedings{bfcl25,
title={The Berkeley Function Calling Leaderboard (BFCL): From Tool Use to Agentic Evaluation of Large Language Models}, 
author={Patil, Shishir G. and Mao, Huanzhi and Cheng-Jie Ji, Charlie and Yan, Fanjia and Suresh, Vishnu and Stoica, Ion and E. Gonzalez, Joseph},
booktitle={Forty-second International Conference on Machine Learning},
year={2025},
}

@inproceedings{basu25,
  title={Nestful: A benchmark for evaluating llms on nested sequences of api calls},
  author={Basu, Kinjal and Abdelaziz, Ibrahim and Kate, Kiran and Agarwal, Mayank and Crouse, Maxwell and Rizk, Yara and Bradford, Kelsey and Munawar, Asim and Kumaravel, Sadhana and Goyal, Saurabh and others},
  booktitle={Proceedings of the 2025 Conference on Empirical Methods in Natural Language Processing},
  pages={33526--33535},
  year={2025}
}

@article{peng2023yarn,
  title={Yarn: Efficient context window extension of large language models},
  author={Peng, Bowen and Quesnelle, Jeffrey and Fan, Honglu and Shippole, Enrico},
  journal={arXiv preprint arXiv:2309.00071},
  year={2023}
}

@misc{zhong25,
      title={ComplexFuncBench: Exploring Multi-Step and Constrained Function Calling under Long-Context Scenario}, 
      author={Lucen Zhong and Zhengxiao Du and Xiaohan Zhang and Haiyi Hu and Jie Tang},
      year={2025},
      eprint={2501.10132},
      archivePrefix={arXiv},
      primaryClass={cs.CL},
}

@article{shao2024deepseekmath,
  title={Deepseekmath: Pushing the limits of mathematical reasoning in open language models},
  author={Shao, Zhihong and Wang, Peiyi and Zhu, Qihao and Xu, Runxin and Song, Junxiao and Bi, Xiao and Zhang, Haowei and Zhang, Mingchuan and Li, YK and Wu, Yang and others},
  journal={arXiv preprint arXiv:2402.03300},
  year={2024}
}

@article{williams1992simple,
  title={Simple statistical gradient-following algorithms for connectionist reinforcement learning},
  author={Williams, Ronald J},
  journal={Machine Learning},
  volume={8},
  number={3-4},
  pages={229--256},
  year={1992}
}

@misc{schulman2017kl,
  title={Approximating KL Divergence},
  author={John Schulman},
  year={2017},
  howpublished={\url{http://joschu.net/blog/kl-approx.html}}
}

@article{yu2025dapo,
  title={DAPO: An Open-Source LLM Reinforcement Learning System at Scale},
  author={Yu, Qiying and Zhang, Zheng and Zhu, Ruofei and Yuan, Yufeng and Zuo, Xiaochen and Yue, Yu and Fan, Tiantian and Liu, Gaohong and Liu, Lingjun and Liu, Xin and Lin, Haibin and Lin, Zhiqi and Ma, Bole and Sheng, Guangming and Tong, Yuxuan and Zhang, Chi and Zhang, Mofan and Zhang, Wang and Zhu, Hang and Zhu, Jinhua and Chen, Jiaze and Chen, Jiangjie and Wang, Chengyi and Yu, Hongli and Dai, Weinan and Song, Yuxuan and Wei, Xiangpeng and Zhou, Hao and Liu, Jingjing and Ma, Wei-Ying and Zhang, Ya-Qin and Yan, Lin and Qiao, Mu and Wu, Yonghui and Wang, Mingxuan},
  journal={arXiv preprint arXiv:2503.14476},
  year={2025}
}

@article{wu2025portool,
  title={PORTool: Tool-Use LLM Training with Rewarded Tree},
  author={Wu, Feijie and Zhu, Weiwu and Zhang, Yuxiang and Chatterjee, Soumya and Zhu, Jiarong and Mo, Fan and Luo, Rodin and Gao, Jing},
  journal={arXiv preprint arXiv:2510.26020},
  year={2025}
}

@inproceedings{zeng2025tool,
  title={Tool Zero: Training Tool-Augmented LLMs via Pure RL from Scratch},
  author={Zeng, Yirong and Ding, Xiao and Hou, Yutai and Wang, Yuxian and Du, Li and Dai, Juyi and Ding, Qiuyang and Tang, Duyu and Tu, Dandan and Liu, Weiwen and others},
  booktitle={Findings of the Association for Computational Linguistics: EMNLP 2025},
  pages={9135--9147},
  year={2025}
}

@inproceedings{ma2024m,
  title={m \& m’s: A benchmark to evaluate tool-use for m ulti-step m ulti-modal tasks},
  author={Ma, Zixian and Huang, Weikai and Zhang, Jieyu and Gupta, Tanmay and Krishna, Ranjay},
  booktitle={European Conference on Computer Vision},
  pages={18--34},
  year={2024},
  organization={Springer}
}
